\documentclass[3p, 11pt]{elsarticle}
\usepackage{lineno,hyperref}
\usepackage{amsfonts}
\usepackage{mathrsfs}
\usepackage{amssymb}
\usepackage[english]{babel}
\usepackage{setspace}
\usepackage{ulem}
\usepackage{lipsum, color}
\usepackage{amsmath}
\usepackage[export]{adjustbox}
\usepackage{natbib}
\usepackage{multirow}
\usepackage{longtable}
\usepackage{float}
\usepackage{adjustbox}
\usepackage{cuted}
\usepackage{float}
\usepackage{chngcntr}
\usepackage{setspace}
\usepackage{isomath}

\numberwithin{equation}{section}

\newtheorem{propos}{Proposition}[section]

\makeatletter
\def\ps@pprintTitle{%
   \let\@oddhead\@empty
   \let\@evenhead\@empty
   \let\@oddfoot\@empty
   \let\@evenfoot\@oddfoot
}
\makeatother

\biboptions{authoryear}

\begin{document}

\begin{frontmatter}
\title{Multiple criteria hierarchy process for sorting problems under uncertainty applied to the evaluation of the operational maturity of research institutions}



\author[tel,rvt,nov]{Renata~Pelissari\corref{cor1}}
\ead{renatapelissari@unicamp.br}

\author[els]{Alvaro~Jos\'e~Abackerli}
\ead{alvaro.abackerli@embrapii.org.br}

\author[tel]{Sarah~Ben~Amor}
\ead{BenAmor@telfer.uottawa.ca}

\author[rvt,mak]{Maria~C\'elia~Oliveira}
\ead{marolivei@unimep.br}

\author[kle]{Kleber~Manoel~Infante}
\ead{kleber.m.infante@gmail.com}

\cortext[cor1]{Corresponding author}

\address[tel]{Telfer Management School, University of Ottawa, Ottawa, Canada}

\address[rvt]{Post Graduate Program of Industrial Engineering, Methodist University of Piracicaba, Santa B\'arbara d'Oeste, SP, Brazil}

\address[nov]{Shool of Applied Sciences, University of Campinas, Limeira, Brazil}

\address[els]{Embrapii-Brazilian Association for Industrial Research and Innovation, Bras\'ilia, DF, Brazil}

\address[mak]{Engineering School, Mackenzie Presbyterian University, S\~ao Paulo, Brazil}

\address[kle]{KMI Software Consulting, Jundia\'i, Brazil}

\begin{abstract}
\small
 Despite the availability of qualified research personnel, up-to-date research facilities and experience in developing applied research and innovation, many worldwide research institutions face difficulties when managing contracted Research and Development (R\&D) projects due to expectations from Industry (private sector), particularly regarding the applied development procedures, managerial processes and timing. Such difficulties have motivated funding agents to create evaluation processes to check whether the operational procedures of funded research institutions are sufficient to provide timely answers to demand for innovation from industry and also to identify aspects that require quality improvement in research development. For this purpose, several multiple criteria decision-making approaches can be applied. In this context, the research institutions are considered as alternatives for funding and their processes for research development as decision criteria. Among the available multiple criteria approaches, sorting methods are one prominent tool to evaluate the operational capacity. However, the first difficulty in applying multiple criteria sorting methods is the need to hierarchically structure multiple criteria in order to represent the intended decision process. Additional challenges include the elicitation of the preference information and the definition of criteria evaluation, since these are frequently affected by some imprecision. In most approaches, all these critical points are neglected, or, at best, only partially considered. In this paper, a new sorting method is proposed to deal with all of those critical points simultaneously. To consider multiple levels for the decision criteria, the FlowSort method is extended to account for hierarchical criteria. To deal with imprecise data, the FlowSort is integrated with fuzzy approaches. To yield solutions that consider fluctuations from imprecise weights, the Stochastic Multicriteria Acceptability Analysis (SMAA) is used. Finally, the proposed method is applied to the evaluation of research institutions, classifying them according to their operational maturity for development of applied research.
\end{abstract}
\small
\begin{keyword}
Preference modeling; Stochastic Multicriteria Acceptability Analysis; Hierarchy criteria; SMAA-FFS; Operational maturity evaluation; Research funding.
\end{keyword}

\end{frontmatter}

\onehalfspacing
\normalsize
\section{Introduction}

    \noindent In Multiple Criteria Decision Aid (or Multiple Criteria Decision Making - MCDA/M), sorting problems refers to sorting decision options into predefined and ordered categories, according to the criteria that characterize the acceptability of decision alternatives. The wide range of real-world sorting problems has constituted the major motivation for researchers in developing methodologies for constructing sorting models \citep{Zopounidis2002}.
    
    Most of the existing MCDA/M methods for solving sorting problems make three basic assumptions: (i) the possibility to define precise performance values of alternatives, as well as values of the parameters required by the models, (ii) the possibility of clear definitions of weights for the decision criteria and (iii) the existence of a finite set of criteria, or, at least, the existence of a small number of criteria not organized hierarchically.
    
    However, as discussed by \cite{Angilella2016}, \cite{CORRENTE20171} and \cite{ARCIDIACONO2018}, real-world decision problems often challenge those assumptions. Therefore, the application of most existing MCDA/M sorting methods is affected by simplifications that neglect several aspects that may impact the final result and consequent decisions. In the last two decades, three important methodological challenges to MCDA/M methods have been discussed: (i) considering multiple types of uncertain and imprecise data in the definition of the performance values of alternatives and of the values of the model' parameters, (ii) indirect elicitation of criteria weights, and (iii) hierarchically structuring of the applied criteria.
    
    In particular, to model uncertain and imprecise data, different sorting methods and integration proposals have been developed, including methods based on probability and fuzzy theories. \cite{Janssen2013171} extended FlowSort using interval theory to cover input data given as interval data. \cite{campos2015} proposed another extension of FlowSort using fuzzy triangular numbers to make it suitable to interval data and linguistic-term modeling. 
    
    Despite these developments, there are methods to model multiple types of uncertain and imprecise data simultaneously. One of them is the SMAA-FFS method proposed by \cite{PELISSARI2019}, which is an extension of Fuzzy-FlowSort based on SMAA that makes it possible to simultaneously model imprecise data and uncertain data. For a survey of techniques to model uncertain and imprecise data in MCDA/M, see \cite{TheodorStewart2012}, \cite{ben-amor2015} and \cite{Pelissari2018}.
    
    Another challenge in developing those methods concerns the assessment of weights for the decision criteria. Most of the MCDA/M methods require a direct elicitation of weights, for which the decision-maker (DM) is supposed to provide clear and deterministic values \citep{VETSCHERA2017244}. Given the complexity of this task, a trend toward indirect elicitation is observed \citep{Angilella2016}. In an indirect elicitation, the DM does not need to provide information regarding preference or, in some cases, the provided preference information can be subjective.
    
    
    There are two well-known MCDA/M methods for the indirect elicitation of preferences, that explore the whole set of preference parameters to represent the DM's possible selection choices: the Robust Ordinal Regression (ROR) \citep{Corrente2013, Corrente2014} and the Stochastic Multi-objective Acceptability Analysis (SMAA) (see \cite{lahdelma-salminen2001} for an introduction to SMAA and \cite{Pelissari2019-SMAA} for a survey of this method). SMAA enables the assignment of ordinal, interval, incomplete or even completely missing weights. In the context of sorting problems, there are four methods based on SMAA: SMAA-OC \citep{Lahdelma2010}, SMAA-TRI \citep{Tervonen2009a}, SMAA integrated with ELECTRE-TRI, and SMAA-FFS, which integrates SMAA with Fuzzy-FlowSort \citep{PELISSARI2019}.
    
    The last challenge, as discussed by \cite{CORRENTE2012660} and \cite{Angilella2016}, is the frequent need to consider hierarchical criteria in real applications. Hierarchy helps to break down complex decision-making problems into manageable sub-tasks, being therefore very attractive to DMs. The Analytic Hierarchy Process (AHP) \citep{saaty1994} method and the Multiple Criteria Hierarchy Process (MCHP) methodology \citep{CORRENTE2012660} are some examples to be considered. While AHP provides recommendations only at the highest level of the hierarchy, the MCHP method provides recommendations at all levels of the hierarchy. \cite{Ishizaka2012} proposed the AHPSort method as an extension of AHP for sorting problems, and \cite{Durbach2014} proposed SMAA-AHP for uncertain data modeling to approach ranking problems instead of sorting problems. As far as the available literature indicates, the only integration of MCHP with sorting methods was proposed by \cite{CORRENTE2016} using the ELECTRE-TRI method, although many integrations have been done with ranking methods \citep{DELVASTOTERRIENTES20154910, CORRENTE2013820, CORRENTE20171}.
    
    The above scenario shows that the methods developed usually account separately for each of the above-mentioned aspects of decision problems. However, there are instances in which these aspects need to be considered together in a single problem (\cite{Angilella2016}). Therefore, although there are methods for sorting problems to deal with imprecise data, uncertain data, elicitation of criteria weights and hierarchically structured criteria, there is still a shortage of sorting methods that consider multiple types of uncertain and imprecise data and when it comes to simultaneously addressing the indirect elicitation of weights and hierarchically structured criteria.
    
    In this paper we undertake all these three challenges jointly. The proposed method is based on an integration of FlowSort and MCHP to consider hierarchically organized criteria. Further, Fuzzy theory is applied to model imprecise data. To obtain solutions which account for the space of fluctuations related to uncertainty/imprecision in criteria weights, Stochastic Multicriteria Acceptability Analysis (SMAA) approach is adopted. The proposed method, which in this paper is designed SMAA-FFS-H, is embedded in an evaluation framework of the operational maturity of research institutions. 
    
    In the context of operational maturity evaluation of research institutions, it may also be important to identify which criteria or sub-criteria should be improved in order to assign an institution to a better category. That information can be obtained using the single-criterion flow concept, which was discussed by \cite{Brans2016} for PROMETHEE. The same idea is used here to calculate the single-criterion flows in the framework of the proposed method.
    
    This paper is organized as follows. Section 2 reviews the FlowSort, Fuzzy-FlowSort and SMAA-FFS methods and also presents the MCHP methodology, which are the foundation of the proposed method. In Section 3, we introduce the new method termed FlowSort-H, which is an extension of FlowSort for hierarchical criteria. Then, in Section 4, we propose the SMAA-FFS-H method, applying fuzzy theory and the SMAA methodology to FlowSort-H. Section 5 presents the real-life case-study of the maturity evaluation of research institutions funded in Brazil. We conclude the paper and present some future research suggestions in Section 6.

\section{Background}\label{section_review}

    \noindent In this section, we present a brief reminder of FlowSort, Fuzzy-FlowSort, SMAA-FFS and MCHP, which are the bases of the method proposed in this paper.
    
    \subsection{FlowSort}\label{sec_flowsort}
  
    \noindent FlowSort is an outranking sorting method proposed by \cite{Nemery200890} based on the PROMETHEE methodology . Considering $A=\{x_1, \ldots, x_m\}$ a set of alternatives evaluated according to a set of criteria $G=\{g_1, g_2, \ldots, g_n\}$, FlowSort aims to assign alternatives of $A$ to $k$ predefined ordered categories $C_1, C_2, \ldots, C_k$, in which $C_1$ is the best category and $C_k$ the worst. The categories in FlowSort can be defined either by a lower and upper limiting profiles or by central profiles. Our discussion will be limited to the case of limiting profiles. 
    
    Let $R = \{r_1, \ldots, r_{k+1}\}$ be the set of reference profiles that delimited the $k$ categories, in which $r_{1}$ and $r_{k+1}$ are the best and the worst reference profiles, respectively. Since the categories are completely ordered, each reference profile is preferred to the successive ones, i.e., the following condition is met:
    
    \begin{equation}\label{condition_profiles} 
         \hbox{Condition:~} r_1 \succ r_2 \succ \ldots \succ r_k \succ r_{k+1}. 
    \end{equation} Moreover, the evaluation of alternatives might be delimited by $r_{1}$ and $r_{k+1}$. 
    
    Let $w_j$ be the weight of importance of criterion $g_j$ such that $w_j > 0$ and $\sum_j w_j = 1$, $\pi(x_i,r_h)$. The outranking degree is given by:
        \begin{equation}
            \pi(x_i, r_h) = \sum_{j=1}^{n}w_j P_j(x_i, r_h),
        \end{equation}\label{outranking_func}
    in which $P_j(x_i, r_h)$ is a preference function, assuming values between 0 and 1. The shape of the preference function for each criterion should be chosen according to the DM preference from six different types as proposed by \cite{Brans2005}.
    
    For any alternative $x_i$, $i=1,\ldots, m$, the set $R_i = R \cup \{x_i\}$ is defined. Considering $x_i \in A$ and $r_h \in R_i, ~i=1, \ldots, m, ~h=1, \ldots, k+1$, the positive, negative, and net flows of the alternative $x_i$ are defined by equations \eqref{positive_flow}, \eqref{negative_flow} and \eqref{net_flow}, respectively: 
        \begin{gather}
            \phi^{+}(x_i) = \frac{1}{|R_i| - 1}\sum_{r_h\in R_{i}} \pi(x_i,r_h), \label{positive_flow}\\
            \phi^{-}(x_i) = \frac{1}{|R_i| - 1}\sum_{r_h\in R_{i}} \pi(r_h,x_i), \label{negative_flow}\\
            \phi(x_i) = \phi^{+} (x_i) - \phi^{-} (x_i). \label{net_flow}
        \end{gather}
    
    \noindent Positive, negative and net flows of the reference profile $r_h \in R_i$ are defined by equations \eqref{positive_flow_profiles}, \eqref{negative_flow_profiles} and \eqref{net_flow_profiles}, respectively:
       \begin{gather}
            \phi_{R_i}^{+}(r_h) = \frac{1}{|R_i| - 1} \left[\sum_{\substack{h=1\\h\neq l}}^{k} \pi(r_h, r_l) + \pi(r_h,x_i) \right], \label{positive_flow_profiles}\\
            \phi_{R_i}^{-}(r_h) = \frac{1}{|R_i| - 1} \left[\sum_{\substack{h=1\\h\neq l}}^{k} \pi(r_l, r_h) + \pi(x_i, r_h) \right],\label{negative_flow_profiles}\\
            \phi_{R_i}(r_h) = \phi_{R_i}^{+}(r_h) - \phi_{R_i}^{-}(r_h). \label{net_flow_profiles}
        \end{gather}

    To assign an alternative to a category, its positive and negative flows are compared to the positive and negative flows of the reference profiles based on the rules presented in \eqref{comp_positive_flow} and \eqref{comp_negative_flow}, respectively:
    \begin{gather}
        \mbox{ if } \phi^{+}_{R_i} (r_h) \geq  \phi^{+} (x_i) > \phi^{+}_{R_i} (r_{h+1}), \mbox{ then } C_{\phi^{+}}(x_i) = C_h, \label{comp_positive_flow}\\
        \mbox{ if } \phi^{-}_{R_i} (r_h) < \phi^{-} (x_i) \leq \phi^{-}_{R_i} (r_{h+1}), \mbox{ then } C_{\phi^{-}}(x_i) = C_h. \label{comp_negative_flow}
    \end{gather}
    
    \noindent In order to assign each alternative to exactly one category, the rule based on net flow presented in \eqref{comp_net_flow} can be used:
    \begin{equation}
        \mbox{ if } \phi_{R_i} (r_h) \geq \phi (x_i) > \phi_{R_i} (r_{h+1}), \mbox{ then } C_{\phi}(x_i) = C_h. \label{comp_net_flow}
    \end{equation}
    
    Therefore, we can see that in FlowSort, after conducting the pairwise comparisons between evaluations of the alternatives and limiting profiles, the flows are calculated. From then on, the evaluations and the values of the limiting profiles are no longer used during the analysis, and the flows are the indicators used for the process of assigning the alternatives in the categories. One of the advantages of using the flows is the fact of having a non-compensatory method, which is a characteristic of outranking methods such as FlowSort. 
    
    \subsection{Fuzzy-FlowSort}
    
    \noindent To allow FlowSort to be applied when evaluation of alternatives are defined by imprecise data, such as linguistic terms, \cite{campos2015} proposed a fuzzy extension of FlowSort, modeling the imprecise data by triangular fuzzy numbers.
    
    A triangular fuzzy number is denoted by $\widetilde{M} = (m; \alpha; \beta)_{LR}$, where $m$ is the mean value of the fuzzy number $\widetilde{M}$, and $\alpha$ and $\beta$ are its left and right boundary values, respectively. A crisp number $w$ can be formulated with $m = w$ and $\alpha=\beta=0$. The necessary algebraic operations to make computations with triangular fuzzy numbers are defined as follows:
    \begin{itemize}
        \item Addition: $\widetilde{M} \oplus \widetilde{N} =(m;\alpha;\beta)_{LR} \oplus (n;\gamma;\delta)_{LR} = (m+n;\alpha+\gamma; \beta + \delta)_{LR}$
        \item Subtraction: $\widetilde{M} \ominus \widetilde{N} = (m;\alpha;\beta)_{LR} \ominus (n;\gamma;\delta)_{LR} = (m - n; \alpha+\delta; \beta + \gamma)_{LR}$
        \item Multiplication by a scalar number: $w \otimes \widetilde{M} = (w,0,0)_{LR} \otimes (m;\alpha;\beta)_{LR} = (w m ;w \alpha;w \beta)_{LR}$
    \end{itemize}
    
    Let be $g_j(x) = (m;\alpha;\beta)_{LR}$ and $g_j(y) = (n;\gamma;\delta)_{LR}$ and $w$ a scalar number. The global fuzzy outranking degree for each pair $(x,y) \in R_i$ can thus be computed as follows:
        \begin{equation}\label{fuzzy_preferences}
        \begin{aligned}
            \widetilde{\pi}(x,y) &= \sum_{j=1}^{m}w_j \ominus \widetilde{P}_j(x,y)\\
                                 &= \sum_{j=1}^{m}w_j \ominus \widetilde{P}_j(\widetilde{g_j}(x) \ominus \widetilde{g_j}(y)).\\
        \end{aligned}
        \end{equation}
    In equation \eqref{fuzzy_preferences}, if the preference function $\widetilde{P}_j(\widetilde{g_j}(x) \ominus \widetilde{g_j}(y))$ is of type V-Shape diffuse (the most used preference function among the six proposed by
    \cite{Brans2005}), it is given by equation \eqref{Pref_func_fuzzy}, as proposed by \cite{Geldermann2000}:
     \begin{equation}\label{Pref_func_fuzzy}
        \begin{aligned}
            \widetilde{P}_j(\widetilde{g_j}(x) &\ominus \widetilde{g_j}(y))  =\\
            & = ~\widetilde{P}_j((m - n ; \alpha+\delta; \beta + \gamma)_{LR})\\
            & = ~(P_j(m - n);\\
            &~~~~~ ~P_j(m - n) - P_j(m - n - \alpha + \delta);\\
            &~~~~~ ~P_j(m - n + \beta + \gamma) - P_j(m- n))_{LR}.
        \end{aligned}
        \end{equation}
        
    Each global fuzzy outranking degree $\widetilde{\pi}(x,y)$ has to be defuzzified, transforming the fuzzy outranking degree into a crisp number. For that, the authors used the Yager’s operator. Therefore, given a triangular fuzzy number $\widetilde{M} = (m; \alpha; \beta)_{LR}$, its defuzzification is given by
    \begin{equation}\label{defuzzification}
        M^{Def} = m + \frac{\beta+\alpha}{3}.
    \end{equation} 
    
    Using the crisp outranking degree obtained from the defuzzified, positive, negative and net flows of each element $x$ of $R_i$ may be computed as in FlowSort and as already presented in equations \eqref{positive_flow}, \eqref{negative_flow} and \eqref{net_flow}. Finally, these crisp flows values may be used to assign each alternative to a category following the traditional FlowSort assignment rules established in equations \eqref{comp_positive_flow}, \eqref{comp_negative_flow} and \eqref{comp_net_flow}. 
    
    \subsection{SMAA-FFS}
    
    \noindent Although Fuzzy-FlowSort can model imprecise data, it cannot model uncertain data and to handle the indirect elicitation of criteria weights. To overcome those limitations, \cite{PELISSARI2019} integrated the SMAA method to Fuzzy-FlowSort. 
    
    In SMAA-FFS, alternatives evaluated by using linguistic terms are modeled by triangular fuzzy numbers, as in Fuzzy-FlowSort. Uncertain or imprecise evaluations of alternatives are represented by random variables $\xi$ with a probability density function $f_{X}(\xi)$ in the space $X$ defined as 
            $$X = \{ \xi \in {\mathbb{R}}^m \times {\mathbb{R}}^n: {\xi}_{ij}, i=1, \ldots, m, j=1, \ldots, n\}.$$
    Different types of information regarding the criteria weights can also be used in SMAA-FFS. Criteria weights may be defined by ordinal weight information, interval weights, completely missing information or partially missing weights information. To represent all of these types of criteria weights (ordinal, interval and missing), SMAA-FFS considers the weights vector $w$ with a probability density function $f_W(w)$ in the feasible weight space $W$. The weights are normalized and non-negative, and therefore the feasible weight space $W$ is given by
        \begin{equation}
            W = \{w \in {\mathbb{R}}^n: w\geq 0 \hbox{ and } \sum_{j=1}^{n}w_{j}=1\}.\label{eq_W}
        \end{equation}
    
    The indifference ($p$) and preference ($q$) thresholds required by some of preference functions can be defined by interval data or stochastic data. In those cases, they are represented by random variables $\tau = (\rho, \eta)$ with a probability function $f_T$ in the space $T$, such that all feasible combinations of thresholds must satisfy $\rho_j - \eta_j \leq 0$ for each criterion $j$, $j=1, \ldots, n$. Therefore, the space $T$ is defined as
                $$T = \{\tau \in \mathbb{R}^n \times \mathbb{R}^n: \tau = (\rho, \eta), \rho_j - \eta_j \leq 0, \forall j=1, \ldots, n\}.$$
                
    SMAA-FFS also permits the use of linguistic terms for limiting profiles. In this case, limiting profiles are represented by triangular fuzzy numbers ($\widetilde{B} = (b; \alpha; \beta)_{LR}$). The upper and lower profiles can also be defined by interval or stochastic data. For those situations, they are represented by random variables $\psi$ with a probability density function $f_{Y}(\psi)$ in the space $Y$ defined by
            $$Y = \{\psi \subseteq \mathbb{R}^{(k+1)} \times \mathbb{R}^n: \psi_{hj} - \psi_{lj} \leq 0, \forall h<l, h=1, \ldots, k, j=1, \ldots, n\}.$$
    As output, SMAA-FFS produces the category acceptability index for all pairs of alternatives and categories. The category acceptability index, denoted by $C_{i}^{h}$, represents the probability of an alternative $x_i$ to be assigned to category $C_h$. 
        
\subsection{Multiple Criteria Hierarchy Process}\label{section_MCHP}

    \noindent The Multiple Criteria Hierarchy Process (MCHP) was proposed by \cite{CORRENTE2012660} for taking into consideration a hierarchical criteria structure in decision-making problems. The basic idea of MCHP is to consider the preference relationships in each node of the hierarchy of the criteria tree. These preference relations concern both the phase of eliciting information about preference and the phase of analyzing a final recommendation by the DM. In MCHP, the following notation is used:

    \begin{itemize}
        \item $\mathscr{G}$ is the set of all criteria at all levels;
        \item $l$ is the number of levels in the criteria hierarchy;
        \item $\mathscr{I}_\mathscr{G}$ is the set of all indexes of all criteria in the hierarchy;
        \item $f$ is the number of criteria of the first level $G_1, \ldots, G_f$;
        \item $G_\mathbf{r}$, with $\mathbf{r} = (j_1, \ldots, j_h) \in \mathscr{I}_\mathscr{G}$, denotes a sub-criterion of the first-level criterion $G_{j_1}$ in level $h$; the first-level criteria are denoted by $G_{j_1}, j_1 = 1, \ldots, f$;
        \item $n(\mathbf{r})$ is the number of sub-criteria of $G_\mathbf{r}$ at the subsequent level, that is, the direct sub-criteria of $G_\mathbf{r}$ are $G_{(\mathbf{r},1)}, \ldots, G_{(\mathbf{r},n(\mathbf{r}))}$;
        \item $g_\mathbf{t}: A \rightarrow \mathbb{R}$, com $\mathbf{t} = (j_1, \ldots, j_l) \in \mathscr{I}_\mathscr{G}$, denotes an elementary sub-criterion of the first-level criterion $G_{j_1}$, that is, a criterion of level $ l $ of the hierarchical tree of $G_{j_1}$;
        \item $EL$ is the set of indices of all elementary sub-criteria:
            \begin{equation}EL = \mathbf{t} = (j_1, \ldots, j_l) \in \mathscr{I}_\mathscr{G} \hbox{ em que ~}  
                    \left\{
                        \begin{array}{l}
                            j_1 = 1, \ldots, f\\
                            j_2 = 1, \ldots, n(j_1)\\
                            \ldots\\
                            j_l = 1, \ldots, n(j_1, \ldots, j_{l-1});\\
                        \end{array}
                    \right.
            \end{equation}
        \item $G_E \subset \mathscr{G}$ is the set of all the elementary criteria in $G$ 
        and $E_{G} \subset \mathscr{I}_\mathscr{G}$ is the set of indexes of the elementary criteria;
        \item $E(G_\mathbf{r})$ is the set of indices of elementary sub-criteria descending from $G_\mathbf{r}$:
            \begin{equation}
            E(G_\mathbf{r}) = \{(\mathbf{r}, j_{h+1}, \ldots, j_l)\} \in \mathscr{I}_\mathscr{G} \hbox{ em que ~}  
                    \left\{
                        \begin{array}{l}
                            j_{h+1} = 1, \ldots, n(\mathbf{r})\\
                            \ldots\\
                            j_l = 1, \ldots, n(\mathbf{r}, j_{h+1}, \ldots, j_{l-1}).\\
                        \end{array}
                    \right.
            \end{equation}
        \noindent Therefore, $E(G_\mathbf{r}) \subset EL.$
    \end{itemize}

    Without loss of generality, we assume that each elementary sub-criteria $g_\mathbf{t}, \mathbf{t} \in EL$, maps alternatives to real numbers $g_\mathbf{t}: A \rightarrow \mathbb{R}$, in such a way that if $g_\mathbf{t}(x) \geq g_\mathbf{t}(y)$ for all $x,y \in A$, it means that $x$ is at least as good as $y$ in relation to the elementary criterion. In MCHP, each alternative $x \in A$ is evaluated with regards to the elementary sub-criteria.

    A minimum requirement that the preference relationship has to satisfy is the dominance principle for the criteria hierarchy, stating that if alternative $x$ is at least as good as alternative $y$ for all sub-criteria $G_{(\mathbf{r}, j)}$ of $G_\mathbf{r}$ from the level immediately below, then $x$ is at least as good as $y$ in $G_\mathbf{r}$. For example, if a student $x$ is at least as good as a student $y$ in algebra and analysis, being these sub-criteria of mathematics, then $x$ is at least as good as $y$ in mathematics.

 \section{FlowSort method extended to the MCHP for hierarchical criteria: the FlowSort-H method}
 
    \noindent In this section, we propose the FlowSort-H method, extending FlowSort to the MCHP. In the FlowSort-H method, criteria are not all located at the same level; instead, they are hierarchically structured. 

    We define the set $\mathscr{G}$ of all criteria in all hierarchical levels, distributed in $l$ different levels. At the first level, there are the root criteria. Each root criterion has its own hierarchical tree with sub-criteria. The leaves of each hierarchy tree are at the last level, which can be different for each root criteria (last level $\leq l$,) and they are called elementary sub-criteria. The alternatives are evaluated at the level of elementary criteria. The mathematical notation used in the FlowSort-H method is similar to that proposed by MCHP, introduced in Section \ref{section_MCHP}.
    
    In mathematical terms, FlowSort-H assigns $m$ alternatives $A = \{x_1, x_2. \ldots, x_m\}$ to $k$ ordered and predefined categories $C_1,C_2,...,C_k$, taking into consideration the performance of the alternatives at the level of elementary criteria. Each category $C_h$ is defined by a lower reference profile $r_h^{\mathbf{t}}$ and a top reference profile $r_{h+1}^{\mathbf{t}}$ with respect to each elementary criterion $g_\mathbf{t}$. A lower limiting profile represents the minimum value that an alternative must have as evaluation in a certain elementary criterion to belong to the respective category. 
    
    Let $R^{\mathbf{t}} = \{r_1^{\mathbf{t}}, r_2^{\mathbf{t}}, \ldots, r_{k+1}^{\mathbf{t}}\}$ be the set of limiting profiles related to criterion $g_\mathbf{t}$. We define the set $R_i^{\mathbf{t}} = R^{\mathbf{t}} \cup \{x_i\}$ as the union of $R^\mathbf{t}$ with alternative $x_i$, for $i=1, \ldots, m$. Since the categories are completely ordered, each reference profile is preferred to the successive ones, i.e., the condition given by
    \begin{equation}
         \hbox{Condition:~} r_1^{\mathbf{t}} \succ  r_2^{\mathbf{t}} \succ \ldots \succ  r_k^{\mathbf{t}} \succ  r_{k+1}^{\mathbf{t}}.\label{condition_profiles_H} 
    \end{equation}
    When categories are defined by limiting profiles, as in the FlowSort method, we must assume that each performance value of each alternative in relation to the criterion $g_\mathbf{t}$ is between $r_{k + 1}^\mathbf{t} $ and $r_1^\mathbf{t}$.

    In order to apply the FlowSort-H method, it is necessary to obtain from the DM, besides the limiting profiles of the categories, some information regarding preference, such as the weights of the criteria and indifference and preference thresholds. We denote by $w_\mathbf{r}$ the weight of the criterion $G_\mathbf{r}$. The criteria weights vector has to be positive and normalized at each hierarchy level for each criterion  $G_\mathbf{r}$. Thus, considering the criteria
    $G_{(\mathbf{r},1)}, \ldots, G_{(\mathbf{r}, n(\mathbf{r}))}$, direct descendant of $G_\mathbf{r}$, $w_{(\mathbf{r},s)} > 0, \forall s=1, \ldots, n(\mathbf{r})$ and $\sum_{s=1}^{n(\mathbf{r})} w_{(\mathbf{r},s)} = 1$. In addition, for each elementary criterion, the DM should define the indifference ($q_\mathbf{t}$) and preference ($p_\mathbf{t}$) thresholds. 
    
    The first step in FlowSort-H is to calculate the outranking degree, as in FlowSort. With only one criterion level, comparisons in FlowSort are made between the alternatives and the reference profiles, calculating the distance $d_j (x_i, r_h) = g_j(x_i) - g_j(r_h) $, for $x_i, r_h \in R_i$. Then, for each criterion, the FlowSort method creates a preference function, $P_j(x_i, r_h) $. The idea remains the same in FlowSort-H. The performance values of the alternatives, defined at the level of the elementary criteria, are compared with the reference profiles of the categories, also defined at the level of the elementary criteria, using the outranking degree function. However, the difference is that the outranking degree is weighted by the weights of the criteria at all levels. This concept of comparison changes the calculation of the outranking degree defined in FlowSort and it is given by:
        \begin{equation}\label{grau_pref_hieraq}
                \pi(x_i, r_h^\mathbf{t}) = \sum_{j_1=1}^{f} \pi_{j_1}(x_i, r_h^\mathbf{t})
        \end{equation} 
    \noindent where 
    \begin{equation}
            \pi_{j_1}(x_i, r_h^\mathbf{t}) = w_{j_1} P_{j_1}(x_i, r_h^\mathbf{t}), \hbox{ in which ~}  
            P_{j_1}(x_i, r_h) = \left\{
                                    \begin{array}{l}
                                        P_{j_1}(x_i, r_h), \hbox{ if~} n(j_1)=0,\\
                                        \displaystyle\sum_{j_2=1}^{n(j_1)}w_{j_2}P_{j_2}(x_i, r_h), \hbox{ if~} n(j_1)\neq0.\\
                                    \end{array}
                                \right.
    \end{equation}
    Analogously, 
    \begin{equation}
                     P_{j_2}(x_i, r_h) = \left\{
                                    \begin{array}{l}
                                        P_{j_2}(x_i, r_h), \hbox{ if~} n(j_2)=0\\
                                        \displaystyle\sum_{j_3=1}^{n(j_2)}w_{j_3}P_{j_3}(x_i, r_h), \hbox{ if~} n(j_2)\neq0\\
                                    \end{array}
                                \right., 
    \end{equation}
    The same logic is applied up to the last level of the tree. Then, the positive and negative flows can be computed by equations \eqref{flow_pos_hier} and \eqref{flow_neg_hier}:
        \begin{gather}
            \phi^{+}(x_i) = \frac{1}{|R_i^{\mathbf{t}}| - 1} \sum_{h=1}^{k} \pi(x_i, r_h^\mathbf{t})\label{flow_pos_hier},\\
            \phi^{-}(x_i) = \frac{1}{|R_i^{\mathbf{t}}| - 1} \sum_{h=1}^{k} \pi(r_h^\mathbf{t}, x_i). \label{flow_neg_hier}
        \end{gather}
    The net flow of alternative $x_i$, as in FlowSort, is defined as the difference between the positive and negative flows:
        \begin{equation}
            \phi(x_i) = \phi^{+}(x_i) - \phi^{-}(x_i).\label{net_flow_hier}
        \end{equation}

    Similarly to FlowSort, in FlowSort-H, to locate the flow of an alternative $x_i$ relative to the reference profiles, flows of limiting profiles related to the alternative $x_i$ must be calculated. The calculation of the flow of a reference profile related to the alternative $x_i$ is based on the comparisons of this profile with the others, and of that profile with the alternative $x_i$, as defined in equations \eqref{eq_fluxo_profile_pos_hier}, \eqref{eq_fluxo_profile_neg_hier} and \eqref{eq_fluxo_profile_total_hier}:
        
        \begin{gather}
            \phi_{R_i^{\mathbf{t}}}^{+}(r_h^\mathbf{t}) = \frac{1}{|R_i^{\mathbf{t}}| - 1} \left[\sum_{\substack{h=1\\h\neq l}}^{k} \pi(r_h^\mathbf{t}, r_l^\mathbf{t}) + \pi(r_h^\mathbf{t},x_i) \right],\label{eq_fluxo_profile_pos_hier}\\
            \phi_{R_i^{\mathbf{t}}}^{-}(r_h^\mathbf{t}) = \frac{1}{|R_i^{\mathbf{t}}| - 1} \left[\sum_{\substack{h=1\\h\neq l}}^{k} \pi(r_l^\mathbf{t}, r_h^\mathbf{t}) + \pi(x_i, r_h^\mathbf{t}) \right],\label{eq_fluxo_profile_neg_hier}\\
            \phi_{R_i^{\mathbf{t}}}(r_h^\mathbf{t}) = \phi_{R_i^{\mathbf{t}}}^{+}(r_h^\mathbf{t}) - \phi_{R_i^{\mathbf{t}}}^{-}(r_h^\mathbf{t}).\label{eq_fluxo_profile_total_hier}
        \end{gather}
    
    Analogous to the hypothesis presented by \cite{Nemery200890}, we have the following preposition:
    \begin{propos}
    The order of the fluxes of the reference profiles is invariant with respect to the alternative $x_i$. Thus, $\forall x_i \in \mathcal{A} $ and $\forall h = 1, \ldots, k + 1$ representing the different limiting profiles, we have
         \begin{gather*}
            \phi_{R_i^{\mathbf{t}}}^{+}(r_h^\mathbf{t}) >  \phi_{R_i^{\mathbf{t}}}^{+}(r_{h+1}^\mathbf{t}),\\
            \phi_{R_i^{\mathbf{t}}}^{-}(r_h^\mathbf{t}) <  \phi_{R_i^{\mathbf{t}}}^{-}(r_{h+1}^\mathbf{t}),\\
            \phi_{R_i^{\mathbf{t}}}(r_h^\mathbf{t}) >  \phi_{R_i^{\mathbf{t}}}(r_{h+1}^\mathbf{t}).
         \end{gather*}
    \end{propos}
    
    This means that, although the values of the fluxes of the limiting profiles depend directly on the alternative $x_i$, their orders always respect the order of the categories. This proposition allows the category $C_h$ to be delimited by the values of the flows of its limiting profiles $\phi_{R_i^{\mathbf{t}}}(r_h^{\mathbf{t}})$ e $\phi_{R_i^{\mathbf{t}}}(r_{h+1}^{\mathbf{t}})$ and is the basis for the FlowSort-H assignment rules, which are similar to the FlowSort's and are presented in equations \eqref{atribuicao_pos_flow}, \eqref{atribuicao_neg_flow} and \eqref{atribuicao_net_flow}:
     \begin{gather}
        \mbox{ if } \phi_{R_i^{\mathbf{t}}}^{+}(r_h^\mathbf{t}) \geq  \phi^{+} (x_i) > \phi_{R_i^{\mathbf{t}}}^{+}(r_{h+1}^\mathbf{t}), \mbox{ then } C_{\phi^{+}}(x_i) = C_h, \label{atribuicao_pos_flow}\\
        \mbox{ if } \phi_{R_i^{\mathbf{t}}}^{+}(r_h^\mathbf{t}) < \phi^{-} (x_i) \leq \phi_{R_i^{\mathbf{t}}}^{+}(r_{h+1}^\mathbf{t}), \mbox{ then } C_{\phi^{-}}(x_i) = C_h, \label{atribuicao_neg_flow}\\
        \mbox{ if } \phi_{R_i^{\mathbf{t}}}^{+}(r_h^\mathbf{t}) \geq \phi (x_i) > \phi_{R_i^{\mathbf{t}}}^{+}(r_{h+1}^\mathbf{t}), \mbox{ then } C_{\phi}(x_i) = C_h. \label{atribuicao_net_flow}
    \end{gather}
    
    For the types of problems that the method proposed here intends to solve, it may also be important to identify which criteria or sub-criteria should be improved in order to assign an alternative to a better category. 
    
    For example, consider a decision problem in which one wishes to identify which of the three aspects of sustainability (social, environmental or economic) should be improved in order to have a more sustainable process. Consider that the criteria are organized hierarchically, with several indicators (second-level sub-criteria) related to each of the social, environmental and economic macro-criteria. Analyzing the information available, one can conclude that the social macro-criterion is the aspect that must be improved in order to have a sustainable process. That information can be obtained using the single-criterion flow concept. However, in order to be able to define improvement actions, it is also necessary to identify which indicators associated with the social aspect are contributing negatively to this aspect being the worst. Thus, it is necessary to obtain decision information not only at the macro-criteria level, but at all levels of the hierarchy. In other words, we shall be able to compute the single-criterion at all levels of the hierarchy.
    
    The concept of single-criterion flow was discussed by \cite{Brans2016} for PROMETHEE and the same idea is used here to calculate the single-criterion flow in FlowSort-H. Given the net flow of alternative $x_i$ defined in equation \eqref{net_flow_hier} and the definition of the outranking degree presented in \eqref{grau_pref_hieraq}, we have
       \begin{eqnarray}
            \phi(x_i) &=& \phi^{+}(x_i) -  \phi^{-}(x_i) = \frac{1}{|R_i^{\mathbf{t}}| - 1} \sum_{h=1}^{k} [\pi(x_i, r_h^\mathbf{t}) -  \pi(r_h^\mathbf{t}, x_i)]\\
             &=& \frac{1}{|R_i^{\mathbf{t}}| - 1} \sum_{h=1}^{k} \sum_{j_1=1}^{f} w_{j_1} [P_{j_1}(x_i, r_h^{\mathbf{t}}) - P_{j_1}(r_h^{\mathbf{t}}, x_i)].\label{fluxo_uncriterio2}
        \end{eqnarray}
    Consequently, 
            \begin{equation}
                \phi(x_i) = \sum_{j_1=1}^{f} w_{j_1}\phi_{j_1}(x_i),
            \end{equation}
        \noindent where $\phi_{j_1}(x_i)$ is the single-criterion flow considering only the criterion $G_\mathbf{j_1}$ and is given by:
            \begin{equation}
                \phi_{j_1}(x_i) =  \frac{1}{|R_i^{\mathbf{t}}| - 1} \sum_{h=1}^{k} \left[P_{j_1}(x_i, r_h^{\mathbf{t}}\right] - P_{j_1}(r_h^{\mathbf{t}}, x_i)].\label{eq_unicriterion_1}
            \end{equation}
     Similarly, the single-criterion flow of alternative $x_i$ considering only the criterion $G_\mathbf{r}$, with 
        $\mathbf{r} = (j_1, \ldots, j_h) \in \mathscr{I}_{\mathscr{G}}$, is given by
            \begin{equation}
                \phi_{\mathbf{r}}(x_i) =  \frac{1}{|R_i^{\mathbf{t}}| - 1} \sum_{h=1}^{k} \left[P_{\mathbf{r}}(x_i, r_h^{\mathbf{t}}\right] - P_{\mathbf{r}}(r_h^{\mathbf{t}}, x_i)],\label{eq_unicriterion_2}
            \end{equation} 
        where
        \begin{equation}\label{eq_p}
                     P_{\mathbf{r}}(x_i, r_h) = \left\{
                                    \begin{array}{l}
                                        P_{\mathbf{r}}(x_i, r_h), \hbox{ if~} n(r)=0\\
                                        \displaystyle\sum_{s=1}^{(\mathbf{r},n(r))}w_{s}P_{s}(x_i, r_h), \hbox{ if~} n(r)\neq0\\
                                    \end{array}
                                \right., 
    \end{equation}
            
    The single-criterion flow of the limiting profile $r_h^{\mathbf{t}}$ considering only the criterion $G_\mathbf{j_1}$ is given by:
            $$ \phi_{R_i^\mathbf{t},j_1}(r_h^{\mathbf{t}}) =  \frac{1}{|R_i| - 1}\left\{
            \sum_{\substack{r_s^{\mathbf{t}} \in R_i^\mathbf{t} \\ r_s^{\mathbf{t}} \neq r_h^{\mathbf{t}}}} \left[P_{j_1}(r_h^{\mathbf{t}}, r_s^{\mathbf{t}}) - P_{j_1}(r_s^{\mathbf{t}}, r_h^{\mathbf{t}})\right] + \left[P_{j_1}(r_h^{\mathbf{t}}, x_i) - P_{j_1}(x_i, r_h^{\mathbf{t}})\right]\right\},$$
        \noindent where $P_{j_1}$ is given such as in \ref{eq_p}.
        
    Single-criterion flows assume values between -1 and 1 and are comparable to the single-criterion flow of the limiting profiles. They can be used to identify assignments of the alternatives based only on one criterion, thus identifying, through the obtained assignment, in which criterion the alternative has a better/ worse performance. Thus, an improvement plan with the objective of assigning the alternative $x_i$ to a higher category may consider prioritizing the improvement of a particular criterion/ sub-criterion over another. To assign an alternative to a specific category based only on one criterion, we propose comparing the single-criterion flow of an alternative to the single-criterion flows of the reference profiles, similarly to the assignment procedure applied in FlowSort (equation \eqref{comp_net_flow}). Therefore, single-criterion assignments are obtained based on the following assignment rule:
    \begin{equation}
        \mbox{ if } \phi_{R_i^\mathbf{t},\mathbf{r}}(r_h^{\mathbf{t}}) \geq \phi_{\mathbf{r}}(x_i) > \phi_{R_i^\mathbf{t},\mathbf{r}}(r_{h+1}^{\mathbf{t}}), \mbox{ then } C_{\phi, \mathbf{r}}(x_i) = C_h. \label{comp_net_flow_H}
    \end{equation}
    
    For a better understanding of the FlowSort-H computation, a numerical example is presented in Appendix A.  
    
\section{Applying Fuzzy and SMAA to FlowSort-H: the proposed method SMAA-FFS-H}

    \noindent While in Section 3 we proposed an extension of the FlowSort for hierarchical criteria, in this section we shall discuss the extension of FlowSort-H to the case with imprecise and uncertain preference information, integrating it to the Fuzzy theory and the SMAA methodology. We call the resulting method SMAA-FFS-H.

    Regarding the application of fuzzy theory to FlowSort-H, we may observe that all concepts, computation steps and outputs of the Fuzzy-FlowSort-H method are analogous to those of the Fuzzy-FlowSort method already presented in Section \ref{section_review}. Therefore, we focus our discussion in this section on the application of the SMAA methodology to the Fuzzy-FlowSort-H method. 
    
    It is worth emphasizing that all concepts related to SMAA-FFS (input data, outputs provided by the method and its algorithm) presented in Section \ref{section_review} are analogously applicable to the SMAA-FFS-H method. Furthermore, the proposed method SMAA-FFS-H introduced in this section can also be seen as an extension of the SMAA-FFS in which, instead of applying the Fuzzy-FlowSort method, the Fuzzy-Flowsort-H method is applied. 
    
    Despite the similarities between SMAA-FFS and SMAA-FFS-H, the SMAA-FFS-H method which we are proposing here gives a new index as output besides the category acceptability index: the single-criterion category acceptability index. That index, denoted by $C_{\mathbf{r}, i}^{h}$, represents the probability of an alternative $x_i$ to be assigned to category $C_h$ when it is analyzed only in regards to criterion $g_\mathbf{r}, \mathbf{r}\in IG$.
    
\begin{figure}[ht!]
            \begin{center}
           \includegraphics[width=16cm]{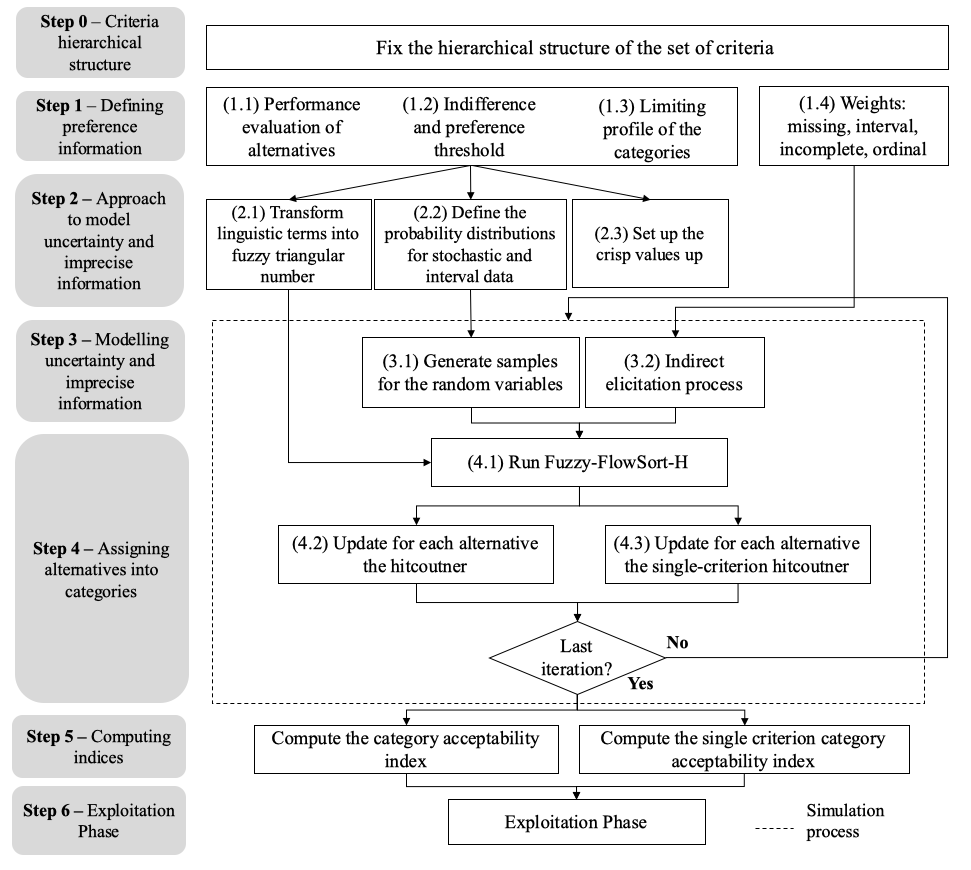}
                \caption{The SMAA-FFS algorithm scheme. Adapted from \cite{PELISSARI2019}.}\label{FlowSort_Scheme}
            \end{center}
        \end{figure}

    The computation of the proposed method SMAA-FFS-H follows the phases showed in Figure \ref{FlowSort_Scheme} and described below. 
        
    \begin{itemize}
        \item \textbf{Step 0} Fix the hierarchical structure of the set of criteria using MCHP, distinguishing the subsets of elementary criteria and higher level criteria, down to the root criterion.
    
        \item \textbf{Step 1} The DM, or the analyst representing the DM, is asked to provide different types of preference information:
        \begin{itemize}
            \item[1.1] Performance evaluation of alternatives has to be defined regarding the elementary-level criteria. 
            \item[1.2] The type of the preference function and indifference and preference threshold has to be defined for each elementary-level criteria.
            \item[1.3] Limiting profile of the categories has to be defined regarding each elementary-level criteria. If $k$ categories are considered, $k+1$ limiting profile need to be defined for each elementary criterion. 
            \item[1.4] Criteria weights may be deterministic, ordinal, intervals, incomplete or even completely missing. 
        \end{itemize}

        \item \textbf{Step 2} Defining approach to model uncertainty and imprecise information:
        \begin{itemize}
            \item[2.1] Preference information defined as linguistic variables has to be transformed into triangular fuzzy numbers. When triangular fuzzy numbers are used to model the evaluations defined by linguistic terms, the pertinence functions can be intercepted. However, to meet the FlowSort-H condition defined in equation \eqref{condition_profiles_H}, the pertinence functions of the limiting profiles cannot be intercepted.
            \item[2.2] For input data defined as random variables, probability distributions have to be defined.
            \item[2.3] If crisp/deterministic values are used, they have to be set. 
            \item[2.4] If non-deterministic weights are defined, a process of indirect elicitation of criteria weights is applied (step 3.2).
        \end{itemize} 
    
        \item \textbf{Step 3} In the third step, the simulation process starts with modeling uncertainty and imprecise preference information:
        \begin{itemize}
            \item[3.1] Firstly, for the input data defined as random variables, random values are generated using their probability distribution (defined at step 2.2). 
            \item[3.2] The process of indirect elicitation of criteria weights is conducted using the algorithms presented by \cite{Tervonen2007a} and also applied in SMAA-FFS, which are described as follows:
            
            - In case of absent preference information, the SMAA method assumes that all weights are equally possible, which is represented by a uniform probability distribution in the feasible weight space $W$ as follows: first, $n-1$ independent random numbers are generated from the uniform distribution in interval [0,1] and sorted in ascending order ($q_1, \ldots, q-1$). Then 0 and 1 are inserted as the first ($ q_0 $) and the last ($q_n$) numbers, respectively. The weights are then obtained as the difference between the consecutive numbers, i.e, $w_j = q_j - q_{j-1}$. 
            
            - The generation of weights from interval or ordinal data is done by restricting the weight space $W$. That space restriction is performed by modifying the weight generation procedure in the absence of information in order to reject weights that do not satisfy the constraints.
            
            - When weights are defined by ordinal data, ordinal preference information is expressed as linear constraints $w_1 \geq w_2 \geq \ldots \geq w_n$. These constraints represent the preference of the DM that the criterion $g_1$ is the most important one, the criterion $g_2$ is the second most important, and so on. It is also possible to consider cases in which the importance of some criterion is not specified or in which equal importance ($w_j = w_k$) is given to two or more criteria. The generation of ordinal weights in the simulation process follows the same logic used in the absence of weight information, except that, after being generated, the weights are sorted according to the ordering established initially.
            
            - In the presence of interval data, the process of generating weights is slightly modified. The upper and lower limits defined for the weights can be implemented by the rejection technique. After a vector of uniformly distributed normalized weights has been generated, the weights are tested at their limits. If any of the constraints are not satisfied, the whole set is rejected, and the weight generation is repeated. 
        \end{itemize} 
        
        \item \textbf{Step 4}  Still into the simulation process, the decision model Fuzzy-FlowSort-H is applied. The result is the assignment of each alternative to a predefined category at all levels of the hierarchy, including single-criterion assignments. Then, the hitcounter in whose category each alternative was assigned by Fuzzy-FlowSort-H is updated (step 4.2). Analogously, the single-criterion hitcounter in whose category each alternative for each criterion (of all hierarchical levels) was assigned by Fuzzy-FlowSort-H is updated (step 4.3). This process might be repeated as many times as iterations are defined. The indicated number of iterations is 10,000 (ten thousand), as presented by \cite{Tervonen2007a}.
            
        \item \textbf{Step 5} After the last iteration and once the simulation process is finished, the category acceptability index is calculated for all pairs of alternatives and categories. The calculation of the category acceptability indices ($C_{i}^{h}$) through simulation is the number of times that the alternative $x_i$ was assigned to the category $C_{h}$ (given by the hitcounter) divided by the number of iterations. The single-criterion category acceptability index is also calculated for all pairs of alternatives and categories, for each criterion of all levels of the hierarchy. The calculation of the single-criterion category acceptability indices ($C_{\mathbf{r}, i}^{h}$) through simulation is the number of times that the alternative ($x_i$) was assigned to the category $C_{h}$ for a specific criterion $g_\mathbf{r}$ (given by the single-criterion hitcounter) divided by the number of iterations. 
        
        \item \textbf{Step 5} The exploitation phase is the last one, in which the DM uses the category acceptability indices to make his/her decision. As proposed by \cite{Lahdelma2010}, this can be made in different ways. If the category acceptability index is equal to 0, the alternative will not be assigned to that category; if it is 1, the alternative will be assigned to that category whatever the combination of the parameter values may be. When an alternative obtains nonzero probabilities for multiple categories, different options to conduct the analysis exist:
            \begin{itemize}
                \item The DM may decide that the current information is not accurate enough to reliably assign the alternatives to categories. In that case, the solution may be to collect more accurate information on the evaluation of alternatives, criteria weights, limiting profiles, thresholds or all of the aforementioned. 
                \item The DM may accept the result that some alternatives are classified into multiple categories.
                \item The DM may classify alternatives based on their category acceptability indices distributions. The DM may assign an alternative to a category whose probability exceeds some threshold, for instance 50\% or another value between 50\% and 100\%. Categories with probabilities close to zero could be excluded (applying another threshold). 
            \end{itemize} 
        \end{itemize}
        
    The SMAA-FFS-H method was implemented in Java.

\section{Case study: operational maturity evaluation of research institutions}

    \noindent In order to show the applicability of the SMAA-FFS-H method and its usefulness in solving real-life decision-making problems, a case study was carried out at the Brazilian Enterprise for Research and Innovation in Industry (EMBRAPII-Empresa Basileira de Pesquisa e Inovacao Industrial in Portuguese) to evaluate the operational maturity of accredited institutions.
    
     To receive financial support from EMBRAPII, non-profitable research institutions must apply for public calls of accreditation and periodically submit operational results to performance evaluations. Off all performance evaluations carried out by EMBRAPII after accreditation, the operational maturity evaluation aims at identifying the strengths and weaknesses of the accredited institutions in the context of Research and Development projects (R\&D) for industry. Details regarding the maturity model to which the SMAA-FFS-H method is applied are discussed in the following sections.

    \subsection{The operational-maturity evaluation}
     
        \noindent The operational-maturity evaluation focuses on nine fundamental processes that are needed for the accredited institutions to operate according to the EMBRAPII business model: project prospecting, technical writing, project negotiation, project management, project execution, portfolio management, intellectual property management, communication and training of human resources.
        
        As observed in many research contexts worldwide, in Brazil most cutting-edge research is carried out at public and private universities. Despite their highly qualified personnel and up-to-date research facilities, it is common for universities to not have a strong business focus or experience when dealing with industries such as research contractors of R\&D projects. 
        
        Although common in industry, the aboved-mentioned processes are key factors to be customized and improved in the academy-industry relationship in order to guarantee full company satisfaction. Those processes also help in the mitigation of risks and contribute to the success of the developed projects. The definition of each process in the present context is found in Table \ref{tab_Process}. 
        
        \begin{table}[ht!]
        \footnotesize
        \caption{Description of the processes assessed in the operational maturity evaluation.}
        \label{tab:processos_embrapii}       
        \begin{tabular}{p{4cm}p{11.5cm}}
        \hline
        Process & Description \\\hline   
        Project Prospecting (PP) & Refers to the activities that seek opportunities to develop R\&D projects for industries, according to the EMBRAPII's business model. \\
        Technical Writing (TW) & Involves the preparation of documents of a technical nature, pertinent to R\&D projects, including but not limited to technical proposals, work plans, contracts, schedules, etc. \\
        Project Negotiation (PN) & Involves the negotiation activities of technical, financial and legal scopes between the accredited institution and the companies interested in the development of R\&D projects.\\
        Project Management (Proj.M) & Related to the typical activities of a project office, including activities scheduling, financial control, management of deadlines, research teams and project milestones, delivery of results to the company, etc., agreed upon in each project between the accredited institution and the company. \\
        Project Execution (PE) & Comprises all the activities inherent to the execution of contacted R\&D projects. \\
        Portfolio Management (Port.M) & Encompasses the consolidated management of all contracted projects, including and not limited to the management of teams in the different projects, the allocation and management of infrastructure, etc., as well as other aspects regarding resource sharing among the current projects and the new development opportunities, within the institution's accredited competencies. \\
        Intellectual Property Management (IPM) & Ranges from supporting the intellectual-property negotiations in contracting the projects (e.g. discussion of IP possibilities, compensation for successful results, royalties, etc.), to the sharing of property rights (e.g. patent co-authorship), support for the drafting and deposit of patent applications, as well as follow-up of results application after the end of the project. \\
        Communication (CM)& Involves the communication initiatives of the accredited institutions, mostly toward private companies aiming at the disclosure of accredited activities, competencies and results of contracted R\&D projects for any audience interested in the accreditation (e.g. the institution itself, companies, public agents, partners, etc.) \\
        Training of Human Resources (THR) & Includes all activities relevant to the hands-on training of human resources for conducting the R\&D projects to industry. \\\hline
        \end{tabular}\label{tab_Process}
        \end{table}    
        
      It is important to point out that, in some cases, academic institutions do perform the activities related to the above-mentioned processes even without having them formally defined. This would be the case, for example, when seeking opportunities for R\&D development without accounting for such activities as a prospection process. Therefore, in the operational-maturity model discussed, the formal setting up of all processes is checked as a first step in the evaluation under the ``existence of processes'' label, followed by the evaluation of their inputs. 
      
      There are five critical process inputs that must be considered for all the above-mentioned processes: Infrastructure, Human Resources, Counterpart, Working Protocols and Institutional References, defined in Table \ref{tab:inputs_embrapii}. 
      
      \begin{table}[ht!]
        \footnotesize
        \caption{Description of the inputs of each process.}
        \label{tab:inputs_embrapii}       
        \begin{tabular}{p{4cm}p{11.5cm}}
        \hline
            Process input & Description \\\hline   
            Infrastructure (IF) & Relates to the physical infrastructure necessary for the execution of the respective process. Thus, for example, the infrastructure for technical writing may consist of offices, computers and software packages; for project management, it may be a typical PMO infrastructure, and for project execution it would be the laboratory facilities themselves. \\
            Human Resources (HR) & All people directly engaged in the activities in all processes (e.g. technical staff and researchers, management and communication team, legal support people, financial and accounting team, etc.) \\
            Counterpart (CR) & Any economic or financial resource required for the contracted R\&D projects, not supplied by EMBRAPII and to be supplied by the accredited institution,  according to the accreditation agreement. \\
            Working Protocols (WP) & Working instructions to be used in the processes, such as operational procedures, working standards and specific rules (e.g. document standards, rules for project management, procurement, customer service, team selection, students selection for training, etc.) \\
            Institutional References (IR) & Institutional policies or local normative instruments to organize and institutionalize the EMBRAPII R\&D activity at the research institution, to guarantee the long-term commitment and the operation consistency within the accreditation. \\\hline
        \end{tabular}
        \end{table} 
      
      Similar to the processes mentioned, the process-inputs are also key-factors to be customized and improved for the proper operation of the EMBRAPII business model. Research facilities (i.e. infrastructure) are usually shared by the universities, and only some of them are accredited by EMBRAPII. In many cases, university rules are sufficient to deal with academic research but not with particular aspects of the EMBRAPII business model, which is organized to create appropriate R\&D services for industry. Finally, EMBRAPII accreditation demands economic or financial contribution from the accredited institution for the contracted R\&D projects, which is not always reflected in university policies. All these operational aspects are accounted for in the maturity model as required inputs for the aboved-mentioned processes, according to the definitions presented in Table \ref{tab:inputs_embrapii}. 
     
    To conduct the operational maturity analysis, three dimensions of processes and the related process-inputs, namely, offer, volume and focus, must be evaluated:
        
         \begin{itemize}
            \item Offer: measures if the processes and their inputs do exist, pointing also to their main provider. Processes or inputs may not be available, may be available/ provided by the accredited institution, or may be available respectively by the research team itself, without formal institutional support. In such cases, the offer dimension would be classified as ``nonexistent,''  ``institution'' or ``research team.'' 
            \item Volume: characterizes the sufficiency or insufficiency of the process or its inputs to meet the accreditation goals. The volume dimension would be characterized as ``sufficient'' or ``insufficient.''
            \item Focus: characterizes the main destination or the priority application of the available processes and their inputs, ranging from processes and inputs fully dedicated to the accredited activities, including projects execution, to processes and inputs barely shared by the R\&D projects or the accredited activities. The focus dimension would be characterized as ``dedicated'' or ``other.''
        \end{itemize}
        
        Figure \ref{fig:model} shows the developed SMAA-FFS-H method applied to the evaluation of the operational maturity of research institutions. The steps for its application are presented in the following sections:
        
     \begin{figure}[ht!]
                \centering
                \includegraphics[width=16cm]{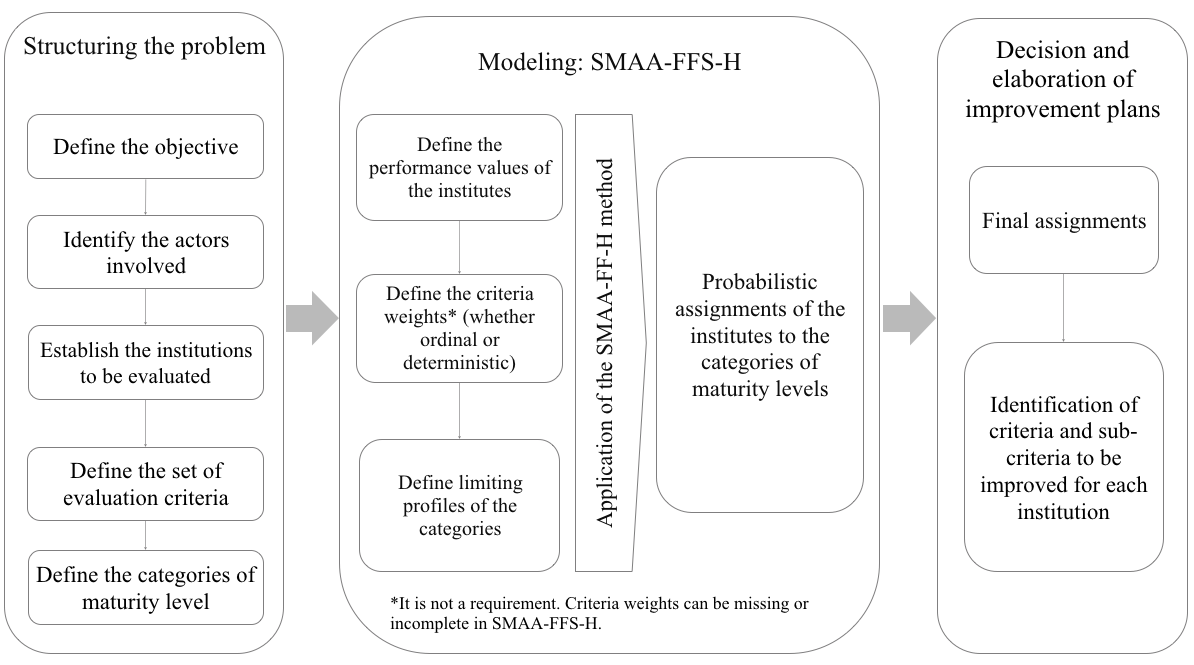}
                \caption{Model to evaluate the operational maturity of research institutions.}
                \label{fig:model}
        \end{figure}

    \subsection{Structuring the problem}

    \noindent The first step of the decision-making process is to establish the goal. In this study, the objective is to classify research institutions into operational maturity levels and to provide a diagnostic of the available operational conditions. Therefore, the problem can be considered as a sorting problem.
    
    The actors involved are the EMBRAPII technicians (i.e. the visitors of the evaluated institutions), the specialist (i.e. a leading technician), the analyst (i.e. the SMAA-FFS-H modeler) and the decision-maker (DM). Technicians are those responsible for the \textit{in loco} evaluation of accredited institutions, collecting operational evidences and answering predefined questions related to the maturity evaluation together with the evaluated institution. Several technicians are involved to cover evaluations all over Brazil. Based on individual answers, a specialist provides a uniform view and a last checkup of the data collected to guarantee the necessary homogeneity in the evaluation. The analyst structures the available information according to the SMAA-FFS-H application. The DM is the COO (Chief Operating Officer) and who defines the decision parameters and rules and makes the final decision.
    
    After defining the objective and identifying the actors involved, the second step is to list the decision alternatives - in this case, the research institutions. In the present case study, eight accredited research institutions are considered. Due to confidentiality reasons, their names have been changed to: Inst. 1, Inst. 2, Inst. 3, Inst. 4, Inst. 5, Inst. 6, Inst. 7, Inst. 8. All institutions were accredited under the same rules and identical accreditation conditions.
    
\begin{figure*}[ht!]
                    \centering
                \includegraphics[width=16cm]{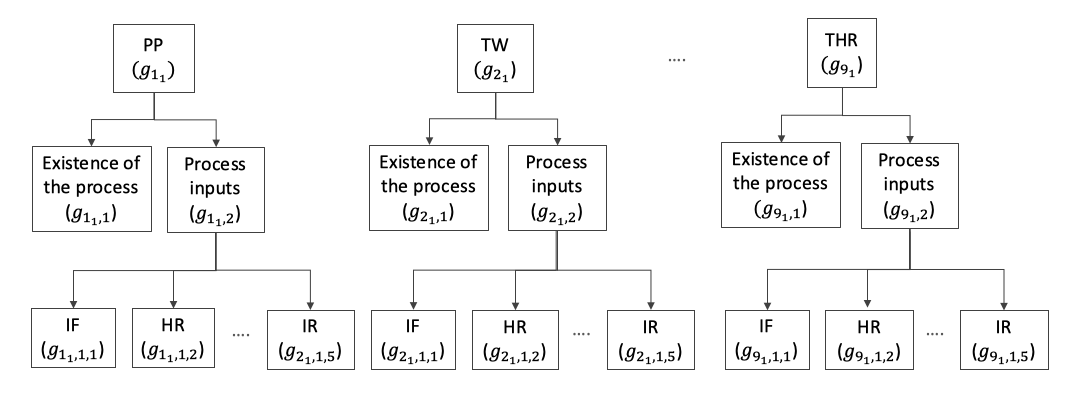}
                    \caption{Hierarchical structure of criteria.}
                    \label{fig:Criteria_Structure}
            \end{figure*}

    The next step is the hierarchical definition of the evaluation criteria. The nine processes listed in Table \ref{tab:processos_embrapii} are the macro-criteria, criteria of the first-level of the hierarchy and are denoted by $g_{1_1}, g_{2_1}, \ldots, g_{9_1}$. Each process has two sub-criteria, ``existence of the process'' and ``process inputs,'' and in turn, the sub-criterion ``process inputs'' has, at the third level of the hierarchy, five sub-criteria, as defined in Table \ref{tab:inputs_embrapii}. For a macro-criterion $g_{j_1}$, $j_1=1,\ldots, 9$, its second-level sub-criteria ``existence of the process'' and ``process inputs'' are denoted by $g_{{j_1},1}$ and $g_{{j_1},2}$, respectively. Each third-level criterion is denoted by $g_{{j_1},2,1}$ $g_{{j_1},2,2}, \ldots, g_{{j_1},2,5}$. The hierarchical structure of criteria is presented in Figure \ref{fig:Criteria_Structure}. Based on this hierarchy, each research institution may be evaluated according to the criterion ``existence of the process'' and to each process inputs (IF, HR, CR, WR, IR).
        
    In order to sort the institutions in levels of operational maturity, it is necessary to define the categories to which they will be assigned; in the present case, four categories or groups are defined as follows:

    \begin{itemize}
        \item Category 1 ($C_1$): Full Maturity -- the institution has complete conditions to comply with the action plan defined at the accreditation.
        \item Category 2 ($C_2$): Consolidated Maturity -- the institution has met a stable and reliable, but not a complete, condition to comply with the accreditation plan.
        \item Category 3 ($C_3$): Structuring Maturity -- the available conditions comply with the accreditation plan but are not stable or reliable yet.
        \item Category 4 ($C_4$): Initial Maturity -- the institution barely complies with the accreditation plan. There are operational conditions already organized, but they are not sufficient to fully fulfill the agreed-upon accreditation plan.
    \end{itemize}

    \subsection{Modeling: application of SMAA-FFS-H}
    
    \noindent After structuring the problem, the modeling phase starts with the assignment of performance values for the institutions, based on a combination of offer, volume and focus for each process and the associated inputs. Table \ref{tab:example_evaluation} shows an example of such an assignment of performance values for the inputs of a process, according to the three dimensions previously mentioned. For each process, the institution performance regarding the criterion ``existence of the process`' might also be evaluated based on a combination of offer, volume and focus.
    
    \begin{table}[ht!]
            \small
            \centering
             \caption{Example of evaluation of the inputs of the process ``Project Prospecting''.}
                \label{tab:example_evaluation}
                \begin{tabular}{llll} \hline
                \multicolumn{4}{c}{Process ``Project Prospecting (PP)''}\\\hline
                Inputs & Offer & Volume & Focus\\\hline
                Infrastructure & Institution & Insufficient & Dedicated \\
                Human Resources & Research team & Sufficient & Other\\
                Counterpart & Nonexistent& & \\
                Working Protocols & Research team & Insufficient & Other\\
                Institutional References  & Nonexistent & & \\\hline
            \end{tabular}
        \end{table}
        
     The evaluation of the offer, volume and focus dimensions is directly related to the level of operational maturity. For example, the combination ``Offer=Institution, Volume=Sufficient and Focus=Dedicated'' represents the highest level of maturity. In turn, when ``Offer=Nonexistent'' and Volume and Focus cannot be assigned (i.e. empty in Table \ref{tab:example_evaluation}), we have the lowest maturity level. Based on these relations, a maturity scale can be defined to be used in the evaluation of each process. See Table \ref{tab:escala_maturidade} for details. 
     
     \begin{table}[ht!]
            \small
            \centering
             \caption{Maturity scale based on combinations of different possible evaluations regarding offer, volume and focus.}
                \label{tab:escala_maturidade}
                \begin{tabular}{cccc} \hline
                Offer    &	Volume	& Focus & Maturity Scale \\\hline
                 Institution         &	Sufficient	      &	Dedicated      &      Extremely Mature (EM)	\\
                 Institution         &	Sufficient	      &	Other         &      Highly Mature (HM)	    \\
                 Institution         &	Insufficient	  &	Dedicated      &      Very Mature (VM) 	    \\
                 Institution         &	Insufficient	  &	Other         &      Slightly Mature (SM)	\\
                 Research team        &	Sufficient	      &	Dedicated      &      Mature (M)	\\
                 Research team        &	Sufficient	      &	Other         &      Slightly Immature (SI)	\\
                 Research team        &	Insufficient	  &	Directed      &      Very Immature (VI)	    \\
                 Research team        &	Insufficient	  &	Other         &      Highly Immature (HI) \\
                 Nonexistent         &	-	              &	-             &      Extremely Immature (EI) \\\hline
            \end{tabular}
        \end{table}
     
\begin{figure*}[htb!]
                \centering
                \small
                \includegraphics[width=\linewidth, frame]{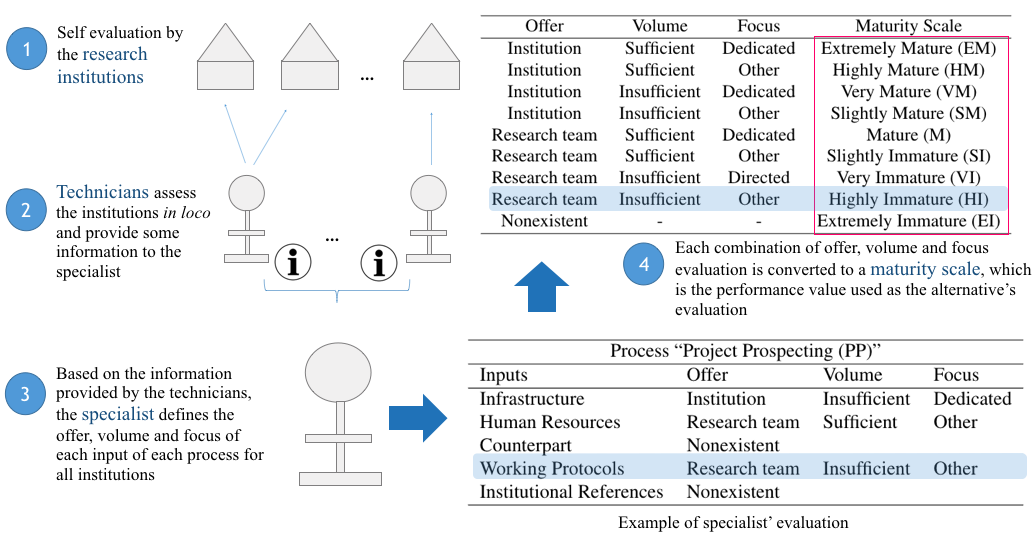}
                \caption{Process to define the performance evaluation of the institutions.}
            \label{fig:Evaluation_Process}
        \end{figure*}

     Considering the above definitions, the operational maturity evaluation starts with a self evaluation by the research institutions, followed by a \textit{in loco} verification of the available operational conditions by EMBRAPII personnel. EMBRAPII technicians collect the available documentation (i.e. formal evidences) and examine the research facilities against the action plan agreed upon at the accreditation. Based on this \textit{in loco} verification and on gathered evidence, the appropriate combination of offer, volume and focus is assigned to each process and to related inputs. A summary of the process of operational maturity evaluation is shown in Figure \ref{fig:Evaluation_Process}.

\begin{figure*}[htb!]
                \centering
                \small
                \includegraphics[width=13cm]{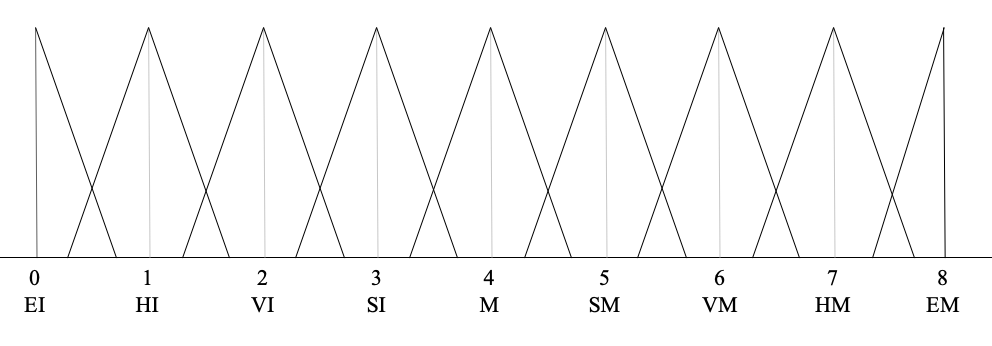}
                \caption{Triangular fuzzy semantic adopted for representing the maturity scale used in the evaluation of alternatives.}
            \label{fig:fuzzy_semantic}
        \end{figure*}

    In the SMAA-FFS-H method, linguistic terms shall be transformed into fuzzy triangular numbers. Table \ref{tab:fuzzy_embrapii} presents the triangular fuzzy semantic used in this study. That semantic is also illustrated in Figure \ref{fig:fuzzy_semantic}.
    
    \begin{table}[ht!]
            \small
            \centering
             \caption{Fuzzy triangular representation used for the maturity scale.}
                \label{tab:fuzzy_embrapii}
                \begin{tabular}{lc} \hline
                 Maturity Scale & Triangular fuzzy semantic\\\hline
                 Extremely Mature (EM) & (8; 0.75; 0)	\\
                 Highly Mature (HM) & (7; 0.75; 0.75)	\\
                 Very Mature (VM)	& (6; 0.75; 0.75)\\
                 Slightly Mature (SM) & (5; 0.75; 0.75)	\\
                 Mature (M) & (4; 0.75; 0.75)	\\
                 Slightly Immature (SI) & (3; 0.75; 0.75)\\
                 Very Immature (VI)	& (2; 0.75; 0.75)\\
                 Highly Immature (HI) & (1;0.75; 0.75)	\\
                 Extremely Immature (EI) & (0; 0; 0.75)	\\\hline
            \end{tabular}
        \end{table}
        
    After setting up the performance evaluation of institutions for operational maturity evaluation, other required parameters must be defined. In order to characterize the categories in which institutions are classified, limiting reference profiles related to each criterion must be defined. In the case of four categories, five limiting profiles are needed. In this case, equal limiting reference profiles were defined for all criteria. The DM wants to make each category narrower than the worst category directly inferior to it. The rationale would be to have progressively narrower categories in the higher maturities, and hence a lower chance of institutions being framed by them. Therefore, the limiting profile values defined were the ones presented in column 3 of Table \ref{tab:perfil_ref_embrapii}.
    
\begin{figure*}[ht!]
                \centering
                \small
                \includegraphics[width=13cm]{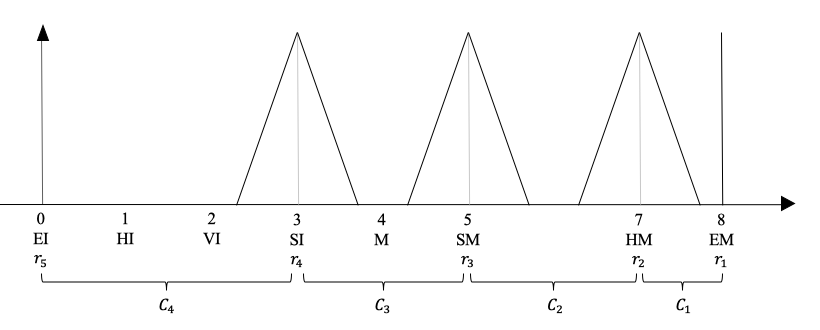}
                \caption{Triangular fuzzy semantic of the limiting profiles of categories.}
            \label{fig:fuzzy_profiles}
        \end{figure*}

    In the SMAA-FFS-H method, reference profiles defined by linguistic terms must also be transformed into triangular fuzzy numbers. Moreover, as required by SMAA-FFS-H, reference profiles can not overlap, $r_1$/$r_5$ has to be greater/less than or equal to the maximum/minimum value of the performance evaluation of the alternatives, and $r_1$ and $r_5$ must be defined by deterministic values. In response to these requirements, the triangular fuzzy semantic presented in column 4 of Table \ref{tab:perfil_ref_embrapii} was adopted. This semantic is also illustrated in Figure \ref{fig:fuzzy_profiles}.
    
    \begin{table}[ht!]
            \small
            \centering
             \caption{Limiting profiles of categories and their fuzzy triangular representation.}
                \label{tab:perfil_ref_embrapii}
                \begin{tabular}{ccccccc} \hline
                 Category & Limiting profile    &	Limiting profile value & Triangular fuzzy semantic\\\hline
                 \multirow{2}{*}{$C_1$} & $r_1$ & Extremely Mature & 8\\
                 \multirow{2}{*}{$C_2$} & $r_2$ & Highly Mature & (7; 0.75; 0.75)\\
                 \multirow{2}{*}{$C_3$} & $r_3$ & Slightly Mature & (5; 0.75; 0.75)\\
                 \multirow{2}{*}{$C_4$} & $r_4$ & Slightly Immature & (3; 0.75; 0.75)\\
                  & $r_5$ & Extremely Immature & 0	\\\hline
            \end{tabular}
        \end{table}
        
    Other parameters usually defined in the SMAA-FFS-H method are the preference and the indifference thresholds. However, when only linguistic variables are used, such as in the present case, these parameters are dispensable. Thus, the preference and the indifference thresholds are zero for all criteria.
    
    When applying the SMAA-FFS-H method, different criteria-weights information can be used (ordinal, deterministic, incomplete, and even missing information). In the present case, we use ordinal preference information. Moreover, the maturity level of the evaluated institutions is prioritized according to the expected longevity of the accredited activities and the available conditions to carry out the projects. Based on this, the DM has set the following preference information: institutional references is the most influential input, infrastructure and human resources, the second most important inputs, followed by counterpart and working protocol inputs, in this order. The priority of the process inputs is presented in Table \ref{tab:weights_input}. 
    
        \begin{table}[ht!]
            \small
            \centering
             \caption{Hierarchical third-level-criteria weights (weights of process inputs, $j=1, \ldots, 9$).}
                \label{tab:weights_input}
                \begin{tabular}{cccccc} \hline
                 Criterion at the third level   & $g_{{j_1},2,1}$& $g_{{j_1},2,2}$& $g_{{j_1},2,3}$& $g_{{j_1},2,4}$& $g_{{j_1},2,5}$\\\hline
                 Ordinal weight (priority) & 2 & 2 & 3 & 4 & 1\\\hline
            \end{tabular}
        \end{table}
    
    In this scenario, the full and effective implementation of individual processes is less important, priority having been given to the projects execution (which requires prospecting and negotiation), followed by the technical writing (so that the contracts, work plans and reports are done) and project and portfolio management in the next priority order (to ensure the completion of contracts). The remaining processes are in the third position in the order of importance. The order of importance of the processes is presented in Table \ref{tab:weights_process}.
    
    \begin{table}[ht!]
            \small
            \centering
             \caption{Hierarchical-first-level-criteria weights (weights of the processes).}
                \label{tab:weights_process}
                \begin{tabular}{cccccccccc} \hline
                 Criterion at the first level   & $g_{1_1}$ & $g_{2_1}$ & $g_{3_1}$ & $g_{4_1}$ & $g_{5_1}$ & $g_{6_1}$ & $g_{7_1}$& $g_{8_1}$& $g_{9_1}$\\\hline
                 Ordinal weight (priority) & 1 & 2 & 1 & 2 & 1 & 2 & 3 & 3 & 3\\\hline
            \end{tabular}
        \end{table}
       
    The last but not least important definition for the application of SMAA-FFS-H to the operational maturity evaluation is the relative importance between the formal existence of processes and their inputs, as per the second level of the decision criteria depicted in Figure \ref{fig:Criteria_Structure}. Such relative importance is given by the weights in Table \ref{tab:weights_second_level}, which indicate a slightly larger importance for the formal existence of the processes when compared with their inputs. 
        
    \begin{table}[ht!]
            \small
            \centering
             \caption{Hierarchical second-level-criteria weights ($j=1, \ldots, 9$.)}
                \label{tab:weights_second_level}
                \begin{tabular}{ccc} \hline
                 Criterion at the second level   & $g_{{j_1},1}$ & $g_{{j_1},2}$\\\hline
                 Interval weight  & 0.6 & 0.4\\\hline
            \end{tabular}
        \end{table}
    
    Once the decision criteria and their importance are defined, the SMAA-FFS-H method can be applied to the evaluated research institutions. We applied the method using the Java code developed. Table \ref{tab:resultado_EMBRAPII_2} displays the category acceptability indices obtained from the application of the SMAA-FFS-H method in the present case, following the steps presented in Figure \ref{FlowSort_Scheme}.
    
       \begin{table}[ht!]
            \small
            \centering
             \caption{Category acceptability index in percentage (\%) with ordinal weights. *Final assignment defined based on the highest acceptability index.}
            \label{tab:resultado_EMBRAPII_2}
            \begin{tabular}{c|ccccc} \hline         
            	&	\multicolumn{4}{c}{Category}			\\\hline  
                Institution	&	$C_1$	&	$C_2$	&	$C_3$	&	$C_4$  & Final assignment*	\\\hline  
                Inst. 1	    &	0	&	0	&	100	&	0	& $C_3$	\\
                Inst. 2	    &	0	&	0	&	0	&	100	& $C_4$	\\
                Inst. 3    	&	0	&	90	&	10	&	0	& $C_2$	\\
                Inst. 4    	&	0	&	0	&	68	&	32	& $C_3$	\\
                Inst. 5    	&	0	&	0	&	0	&	100	& $C_4$	\\
                Inst. 6    	&	0	&	0	&	0	&	100	& $C_4$	\\
                Inst. 7    	&	0	&	0	&	0	&	100	& $C_4$	\\
                Inst. 8     	&	0	&	0	&	0	&	100	& $C_4$	\\\hline
            \end{tabular}
        \end{table}
    
    The results presented in Table \ref{tab:resultado_EMBRAPII_2} show that no institution has full maturity and, therefore, no institution has met all the conditions to fully comply with the accreditation plan. Institution Inst. 3 shows consolidated maturity and, therefore, has stable and reliable operational conditions which are not sufficient for long-term running. Institution Inst. 1 and Inst. 4 show structuring maturity (i.e. were assigned to category $C_3$), with operational conditions not stable or reliable yet. Finally, institution Inst. 2, Inst. 5, Inst. 6, Inst. 7 and Inst. 8, which were assigned to category $C_4$, barely comply with their action plans due to their operational conditions.
    
    \subsection{Decision and elaboration of improvement plans}
    
    \noindent From the results presented in Table \ref{tab:resultado_EMBRAPII_2}, one can see that the eight evaluated institutions must implement improvements to increase the observed operational maturity levels, in order to fully comply with individual accreditation plans. To help identify opportunities for improvements, single-criterion flows must be evaluated to point out the sub-criteria (processes and inputs) of lower maturity levels, which are likely the most influential factors in achieving maturity. The assignments of the institutions, based on the single-criterion flows of the first-level criteria, ``Process'', are presented in Table \ref{tab:resultado_EMBRAPII_3}. 
    
    \begin{table}[ht!]
            \small
            \centering
             \caption{Assignments of the institutions based on first-level-single-criterion flows. *Final assignment defined based on the highest acceptability index.}
            \label{tab:resultado_EMBRAPII_3}
            \begin{tabular}{|l|c|c|c|c|c|c|c|c|} \hline         
                	 &  \multicolumn{8}{|c|}{Final assignment*}  	\\\hline  
                      Process                  & Inst. 1	  &	Inst. 2 &	Inst. 3 &	Inst. 4 &	Inst. 5 &	Inst. 6 &	Inst. 7& Inst. 8 \\\hline
                Project prospecting	&	$C_4$	&	$C_4$	&	$C_2$	&	$C_3$	&	$C_4$	&	$C_4$	&	$C_4$	&	$C_4$	\\
                Technical writing	&	$C_2$	&	$C_3$	&	$C_3$	&	$C_3$	&	$C_4$	&	$C_4$	&	$C_4$	&	$C_4$	\\
                Project negotiation	&	$C_3$	&	$C_3$	&	$C_2$	&	$C_3$	&	$C_4$	&	$C_4$	&	$C_4$	&	$C_4$	\\
                Project management	&	$C_2$	&	$C_4$	&	$C_2$	&	$C_3$	&	$C_4$	&	$C_4$	&	$C_4$	&	$C_4$	\\
                Project execution	&	$C_2$	&	$C_3$	&	$C_3$	&	$C_4$	&	$C_4$	&	$C_3$	&	$C_4$	&	$C_3$	\\
                Portfolio management	&	$C_4$	&	$C_4$	&	$C_3$	&	$C_4$	&	$C_4$	&	$C_4$	&	$C_4$	&	$C_4$	\\
                Intellectual property management	&	$C_2$	&	$C_4$	&	$C_4$	&	$C_4$	&	$C_2$	&	$C_4$	&	$C_4$	&	$C_4$	\\
                Communication	&	$C_4$	&	$C_4$	&	$C_4$	&	$C_4$	&	$C_4$	&	$C_4$	&	$C_4$	&	$C_4$	\\
                Training of human resources	&	$C_2$	&	$C_4$	&	$C_2$	&	$C_2$	&	$C_4$	&	$C_4$	&	$C_4$	&	$C_4$	\\\hline
                
            \end{tabular}
        \end{table}
        
    Here an explanation about the construction of improvement plans for institutions Inst. 1 and Inst. 3 is given, analyzing the single-criterion flows as an example. For the remaining institutions, improvement plans may be similarly obtained but will be omitted here due for lack of space.
    
    Let us begin by analyzing the single-criterion flows of the first-level criteria of Inst. 1 (see Table \ref{tab:resultado_EMBRAPII_3}). Inst. 1 has reached consolidated maturity level ($C_2$) in technical writing, project management, project execution, intellectual property management and training of human resources. In the processes project prospecting, portfolio management and communication, Inst. 1 reached an initial maturity level ($C_4$). Therefore, in the case of Inst. 1, it would be appropriate to implement primarily improvement actions related to these three last processes.
    
    To effectively implement improvement actions related to the processes, single-criterion flows of the second-level criteria must be analyzed. That is, for Inst. 1, we aim to identify whether the main shortcoming of project prospecting, portfolio management and communication concerns the ``existence of the process'' or the ``inputs,'' as shown, as shown in Table \ref{tab:resultado_EMBRAPII_4}. From Table \ref{tab:resultado_EMBRAPII_4}, one can see that the ``existence of the process'' as well the ``process inputs'' for all the three processes, is categorized as Initial Maturity ($C_4$). Hence, priority improvement actions should be conducted in both directions, improving the formal implementation of the three processes and their inputs. Thus, the next step is to identify which inputs should first be improved, which requires analyzing the single-criterion flows of the third-level (L3), also shown in Table \ref{tab:resultado_EMBRAPII_4}.

    \begin{table}[ht!]
            \small
            \centering
             \caption{Assignments of Inst. 1 based on single-criterion flows of second and third levels regarding processes of project prospecting, portfolio management and communication (L2-second level, L3-third level.) *Final assignment defined based on the highest acceptability index.}
            \label{tab:resultado_EMBRAPII_4}
            \begin{tabular}{|l|c|} \hline         
                Process/ Input	 &  Final assignment* 	\\\hline  
                Project prospecting        &  	\\
                ~~~~~L2 - Existence of the process &  $C_4$    \\
                ~~~~~L2 - Process inputs           &  $C_4$    \\\
                ~~~~~~~~~L3 - Infrastructure               &  $C_4$   \\
                ~~~~~~~~~~L3 - Human resources              &  $C_4$   \\	
                ~~~~~~~~~~L3 - Counterpart                  &  $C_4$  \\
                ~~~~~~~~~~L3 - Working Protocols	        &  $C_4$   \\
                ~~~~~~~~~~L3 - Institutional References     &  $C_4$ \\\hline
                Portfolio management & 		\\
                ~~~~~L2 - Existence of the process &  $C_4$    \\
                ~~~~~L2 - Process inputs           &  $C_4$    \\\
                ~~~~~~~~~L3 - Infrastructure               &  $C_4$   \\
                ~~~~~~~~~~L3 - Human resources              &  $C_4$   \\	
                ~~~~~~~~~~L3 - Counterpart                  &  $C_4$  \\
                ~~~~~~~~~~L3 - Working Protocols	        &  $C_4$   \\
                ~~~~~~~~~~L3 - Institutional References     &  $C_4$ \\\hline
                Communication & 		\\
                ~~~~~L2 - Existence of the process &  $C_4$    \\
                ~~~~~L2 - Process inputs           &  $C_4$    \\\
                ~~~~~~~~~L3 - Infrastructure               &  $C_4$   \\
                ~~~~~~~~~~L3 - Human resources              &  $C_4$   \\	
                ~~~~~~~~~~L3 - Counterpart                  &  $C_4$  \\
                ~~~~~~~~~~L3 - Working Protocols	        &  $C_4$   \\
                ~~~~~~~~~~L3 - Institutional References     &  $C_4$ \\\hline
            \end{tabular}
        \end{table}
    
    Such analysis shows initial maturity in all the inputs of all three processes. Thus, the single-criterion flows of the inputs are not sufficient to define a prioritization, as all lead to a categorization of Inst. 1 at the worst maturity level.
        
    Analyzing Inst. 3, it can be seen in Table \ref{tab:resultado_EMBRAPII_3} that Intellectual Property Management and Communication are the least mature processes of this institution. In Table \ref{tab:resultado_EMBRAPII_5}, the maturity of these two processes is detailed, focusing on the second and third levels and pointing out the ``existence of the process'' and some of their inputs in the lower maturity level ($C_4$). This might indicate the need for improvement. Subsequent improvement actions may be focused on the improvement of process inputs that lead Inst. 3 to be assigned to the $C_3$-maturity level in both processes.
    
    \begin{table}[ht!]
            \small
            \centering
             \caption{Assignments of Inst. 3 based on single-criterion flows of second and third levels regarding processes of intellectual property management and communication (L2-second level, L3-third level.) *Final assignment defined based on the highest acceptability index.}
            \label{tab:resultado_EMBRAPII_5}
            \begin{tabular}{|l|c|} \hline         
                Process/ Input	 &  Final assignment* 	\\\hline  
                Intellectual property management       &  	\\
                ~~~~~L2 - Existence of the process &  $C_4$    \\
                ~~~~~L2 - Process inputs           &  $C_3$    \\\
                ~~~~~~~~~L3 - Infrastructure               &  $C_3$   \\
                ~~~~~~~~~~L3 - Human resources              &  $C_3$   \\	
                ~~~~~~~~~~L3 - Counterpart                  &  $C_3$  \\
                ~~~~~~~~~~L3 - Working Protocols	        &  $C_2$   \\
                ~~~~~~~~~~L3 - Institutional References     &  $C_4$ \\\hline
                Communication & 		\\
                ~~~~~L2 - Existence of the process &  $C_4$    \\
                ~~~~~L2 - Process inputs           &  $C_3$    \\\
                ~~~~~~~~~L3 - Infrastructure               &  $C_4$   \\
                ~~~~~~~~~~L3 - Human resources              &  $C_3$   \\	
                ~~~~~~~~~~L3 - Counterpart                  &  $C_3$  \\
                ~~~~~~~~~~L3 - Working Protocols	        &  $C_4$   \\
                ~~~~~~~~~~L3 - Institutional References     &  $C_4$ \\\hline
            \end{tabular}
        \end{table}
        
    Following the same reasoning of analysis, different action plans can be defined to improve the operational maturity level of each evaluated institution.

\section{Conclusions}

\noindent In this paper, the SMAA-FFS-H method was presented as a novel method for sorting problems with interacting criteria. The proposed method is an integration of SMAA and Fuzzy theory with FlowSort-H, the latter also developed here. FlowSort-H is an extension of FlowSort for hierarchical criteria based on the Multiple Criteria Hierarchy Process (MCHP). Based on its configuration, SMAA-FFS-H can be seen as an extension of the existent method SMAA-FFS for hierarchical criteria, using the FlowSort-H method as a replacement for the FlowSort method.

Similar to SMAA-FFS, SMAA-FFS-H is able to model imperfect data and to deal with indirect elicitation of criteria weights. However, unlike SMAA-FFS, the SMAA-FFS-H method allows criteria to be hierarchically organized. Another characteristic of the proposed method is the possibility of identifying which criteria or sub-criteria should be improved (through the single-criterion flows) in order to increase the overall result, providing decision information at all levels of the hierarchy.

 The innovative contribution of this paper can be summarized in three main points: 
    \begin{itemize}
        \item proposal of FlowSort-H as an extension of FlowSort to criteria that are organized hierarchically;
        \item proposal of SMAA-FFS-H to simultaneously consider criteria that are hierarchically organized and to take into account different types of imperfect data concerning the weights and alternative evaluations;
        \item the availability of single-criterion information at all levels of the hierarchy, as an additional output from the SMAA-FFS-H method.
    \end{itemize}


The proposed method was applied to a real-life case-study of evaluating the operational maturity evaluation of research institutions. Eight institutions were sorted in four categories of level of maturity. The maturity was successfully evaluated and priorities for an action plan were defined to increase the overall operational maturity of the evaluated institutions.

A direction for future research is the application of preference desegregation procedures to SMAA-FFS-H in order to estimate the limiting profiles of the categories, requiring only decision examples from the DM, which may be easier than the definition of the limiting profiles themselves. In the case study conducted here, we could also identify the need to consider interaction between criteria. Therefore, another possibility for future work is to extend the SMAA-FFS-H method to consider interactions among criteria, for instance applying the Choquet integral preference model.

\section*{Acknowledgments}
   This study was funded by the Coordination of Improvement of Higher Education Personnel-Brazil (CAPES)-Finance Code 001 and by the S\~ao Paulo Research Foundation (FAPESP)-process number 2018/23447-4.

\appendix
\section*{References}
\renewcommand{\bibname}{Bibliografia} 
\addcontentsline{toc}{chapter}{Bibliografia}
\bibliography{refs}       

\begin{thebibliography}{30}
\expandafter\ifx\csname natexlab\endcsname\relax\def\natexlab#1{#1}\fi
\providecommand{\url}[1]{\texttt{#1}}
\providecommand{\href}[2]{#2}
\providecommand{\path}[1]{#1}
\providecommand{\DOIprefix}{doi:}
\providecommand{\ArXivprefix}{arXiv:}
\providecommand{\URLprefix}{URL: }
\providecommand{\Pubmedprefix}{pmid:}
\providecommand{\doi}[1]{\href{http://dx.doi.org/#1}{\path{#1}}}
\providecommand{\Pubmed}[1]{\href{pmid:#1}{\path{#1}}}
\providecommand{\bibinfo}[2]{#2}
\ifx\xfnm\relax \def\xfnm[#1]{\unskip,\space#1}\fi
\bibitem[{Angilella et~al.(2016)Angilella, Corrente, Greco \&
  S{\l}owi{\'{n}}ski}]{Angilella2016}
\bibinfo{author}{Angilella, S.}, \bibinfo{author}{Corrente, S.},
  \bibinfo{author}{Greco, S.}, \& \bibinfo{author}{S{\l}owi{\'{n}}ski, R.}
  (\bibinfo{year}{2016}).
\newblock \bibinfo{title}{{Robust Ordinal Regression and Stochastic
  Multiobjective Acceptability Analysis in multiple criteria hierarchy process
  for the Choquet integral preference model}}.
\newblock {\it \bibinfo{journal}{Omega}\/},  {\it \bibinfo{volume}{63}\/},
  \bibinfo{pages}{154--169}.
\bibitem[{Arcidiacono et~al.(2018)Arcidiacono, Corrente \&
  Greco}]{ARCIDIACONO2018}
\bibinfo{author}{Arcidiacono, S.~G.}, \bibinfo{author}{Corrente, S.}, \&
  \bibinfo{author}{Greco, S.} (\bibinfo{year}{2018}).
\newblock \bibinfo{title}{Gaia-smaa-promethee for a hierarchy of interacting
  criteria}.
\newblock {\it \bibinfo{journal}{European Journal of Operational Research}\/},
  {\it \bibinfo{volume}{270}\/}, \bibinfo{pages}{606 -- 624}.
\bibitem[{Ben~Amor et~al.(2015)Ben~Amor, Martel \& Guitouni}]{ben-amor2015}
\bibinfo{author}{Ben~Amor, S.}, \bibinfo{author}{Martel, J.~M.}, \&
  \bibinfo{author}{Guitouni, A.} (\bibinfo{year}{2015}).
\newblock \bibinfo{title}{A synthesis of information imperfection
  representations for decision aid}.
\newblock {\it \bibinfo{journal}{Information Systems and Operational
  Research}\/},  {\it \bibinfo{volume}{53}\/}, \bibinfo{pages}{68--77}.
\bibitem[{Brans et~al.(1986)Brans, Vincke \& Mareschal}]{brans1986}
\bibinfo{author}{Brans, J.}, \bibinfo{author}{Vincke, P.}, \&
  \bibinfo{author}{Mareschal, B.} (\bibinfo{year}{1986}).
\newblock \bibinfo{title}{{How to select and how to rank projects: The
  Promethee method}}.
\newblock {\it \bibinfo{journal}{European Journal of Operational Research}\/},
  {\it \bibinfo{volume}{24}\/}, \bibinfo{pages}{228--238}.
\bibitem[{Brans \& Mareschal(2005)}]{Brans2005}
\bibinfo{author}{Brans, J.-P.}, \& \bibinfo{author}{Mareschal, B.}
  (\bibinfo{year}{2005}).
\newblock \bibinfo{title}{Promethee methods}.
\newblock In {\it \bibinfo{booktitle}{Multiple Criteria Decision Analysis:
  State of the Art Surveys}\/} (pp. \bibinfo{pages}{163--186}).
\newblock \bibinfo{address}{New York, NY}: \bibinfo{publisher}{Springer New
  York}.
\bibitem[{Brans \& Smet(2016)}]{Brans2016}
\bibinfo{author}{Brans, J.-P.}, \& \bibinfo{author}{Smet, Y. D. S.~D.}
  (\bibinfo{year}{2016}).
\newblock \bibinfo{title}{{PROMETHEE Methods}}.
\newblock In \bibinfo{editor}{S.~Greco}, \bibinfo{editor}{M.~Ehrgott}, \&
  \bibinfo{editor}{J.~R. Figueira} (Eds.), {\it \bibinfo{booktitle}{Multiple
  Criteria Decision Analysis - State of the Art Surveys}\/} (pp.
  \bibinfo{pages}{187--219}).
\newblock \bibinfo{address}{Berlin, Heidelberg}:
  \bibinfo{publisher}{Springer-Verlag New York}.
\bibitem[{Campos et~al.(2015)Campos, Mareschal \& de~Almeida}]{campos2015}
\bibinfo{author}{Campos, A. C. S.~M.}, \bibinfo{author}{Mareschal, B.}, \&
  \bibinfo{author}{de~Almeida, A.~T.} (\bibinfo{year}{2015}).
\newblock \bibinfo{title}{{Fuzzy Flow Sort: An integration of the FlowSort
  method and Fuzzy Set Theory for decision making on the basis of inaccurate
  quantitative data}}.
\newblock {\it \bibinfo{journal}{Information Sciences}\/},  {\it
  \bibinfo{volume}{293}\/}, \bibinfo{pages}{115--124}.
\bibitem[{Corrente et~al.(2017)Corrente, Figueira, Greco \&
  Słowiński}]{CORRENTE20171}
\bibinfo{author}{Corrente, S.}, \bibinfo{author}{Figueira, J.~R.},
  \bibinfo{author}{Greco, S.}, \& \bibinfo{author}{Słowiński, R.}
  (\bibinfo{year}{2017}).
\newblock \bibinfo{title}{{A robust ranking method extending ELECTRE III to
  hierarchy of interacting criteria, imprecise weights and stochastic
  analysis}}.
\newblock {\it \bibinfo{journal}{Omega}\/},  {\it \bibinfo{volume}{73}\/},
  \bibinfo{pages}{1 -- 17}.
\bibitem[{Corrente et~al.(2014)Corrente, Greco, Kadzinski \&
  Słowinski}]{Corrente2014}
\bibinfo{author}{Corrente, S.}, \bibinfo{author}{Greco, S.},
  \bibinfo{author}{Kadzinski, M.}, \& \bibinfo{author}{Słowinski, R.}
  (\bibinfo{year}{2014}).
\newblock \bibinfo{title}{Robust ordinal regression}.
\newblock {\it \bibinfo{journal}{Wiley Encyclopedia of Operations Research and
  Management Science}\/},  (pp. \bibinfo{pages}{1--10}).
\bibitem[{Corrente et~al.(2016)Corrente, Greco \&
  S{\l}{\'{n}}ski}]{CORRENTE2016}
\bibinfo{author}{Corrente, S.}, \bibinfo{author}{Greco, S.}, \&
  \bibinfo{author}{S{\l}{\'{n}}ski, R.} (\bibinfo{year}{2016}).
\newblock \bibinfo{title}{{Multiple Criteria Hierarchy Process for ELECTRE Tri
  methods}}.
\newblock {\it \bibinfo{journal}{European Journal of Operational Research}\/},
  {\it \bibinfo{volume}{252}\/}, \bibinfo{pages}{191 -- 203}.
\bibitem[{Corrente et~al.(2012)Corrente, Greco \&
  S{\l}owi{\'{n}}ski}]{CORRENTE2012660}
\bibinfo{author}{Corrente, S.}, \bibinfo{author}{Greco, S.}, \&
  \bibinfo{author}{S{\l}owi{\'{n}}ski, R.} (\bibinfo{year}{2012}).
\newblock \bibinfo{title}{{Multiple Criteria Hierarchy Process in Robust
  Ordinal Regression}}.
\newblock {\it \bibinfo{journal}{Decision Support Systems}\/},  {\it
  \bibinfo{volume}{53}\/}, \bibinfo{pages}{660 -- 674}.
\bibitem[{Corrente et~al.(2013)Corrente, Greco \&
  Słowiński}]{CORRENTE2013820}
\bibinfo{author}{Corrente, S.}, \bibinfo{author}{Greco, S.}, \&
  \bibinfo{author}{Słowiński, R.} (\bibinfo{year}{2013}).
\newblock \bibinfo{title}{{Multiple Criteria Hierarchy Process with ELECTRE and
  PROMETHEE}}.
\newblock {\it \bibinfo{journal}{Omega}\/},  {\it \bibinfo{volume}{41}\/},
  \bibinfo{pages}{820 -- 846}.
\bibitem[{Durbach et~al.(2014)Durbach, Lahdelma \& Salminen}]{Durbach2014}
\bibinfo{author}{Durbach, I.}, \bibinfo{author}{Lahdelma, R.}, \&
  \bibinfo{author}{Salminen, P.} (\bibinfo{year}{2014}).
\newblock \bibinfo{title}{{The analytic hierarchy process with stochastic
  judgements}}.
\newblock {\it \bibinfo{journal}{European Journal of Operational Research}\/},
  {\it \bibinfo{volume}{238}\/}, \bibinfo{pages}{552--559}.
\bibitem[{Durbach \& Stewart(2012)}]{TheodorStewart2012}
\bibinfo{author}{Durbach, I.~N.}, \& \bibinfo{author}{Stewart, T.~J.}
  (\bibinfo{year}{2012}).
\newblock \bibinfo{title}{Modeling uncertainty in multi-criteria decision
  analysis}.
\newblock {\it \bibinfo{journal}{European Journal of Operational Research}\/},
  {\it \bibinfo{volume}{223}\/}, \bibinfo{pages}{1--14}.
\bibitem[{Geldermann et~al.(2000)Geldermann, Spengler \&
  Rentz}]{Geldermann2000}
\bibinfo{author}{Geldermann, J.}, \bibinfo{author}{Spengler, T.}, \&
  \bibinfo{author}{Rentz, O.} (\bibinfo{year}{2000}).
\newblock \bibinfo{title}{Fuzzy outranking for environmental assessment. case
  study: iron and steel making industry}.
\newblock {\it \bibinfo{journal}{Fuzzy Sets and Systems}\/},  {\it
  \bibinfo{volume}{115}\/}, \bibinfo{pages}{45--65}.
\bibitem[{Greco et~al.(2013)Greco, Mousseau \& Słowinski}]{Corrente2013}
\bibinfo{author}{Greco, S.}, \bibinfo{author}{Mousseau, V.}, \&
  \bibinfo{author}{Słowinski, R.} (\bibinfo{year}{2013}).
\newblock \bibinfo{title}{Robust ordinal regression in preference learning and
  ranking}.
\newblock {\it \bibinfo{journal}{Machine Learning}\/},  {\it
  \bibinfo{volume}{93}\/}, \bibinfo{pages}{381--422}.
\bibitem[{Ishizaka et~al.(2012)Ishizaka, Pearman \& Nemery}]{Ishizaka2012}
\bibinfo{author}{Ishizaka, A.}, \bibinfo{author}{Pearman, C.}, \&
  \bibinfo{author}{Nemery, P.} (\bibinfo{year}{2012}).
\newblock \bibinfo{title}{Ahpsort: an ahp-based method for sorting problems}.
\newblock {\it \bibinfo{journal}{International Journal of Production
  Research}\/},  {\it \bibinfo{volume}{50}\/}, \bibinfo{pages}{4767--4784}.
\bibitem[{Janssen \& Nemery(2013)}]{Janssen2013171}
\bibinfo{author}{Janssen, P.}, \& \bibinfo{author}{Nemery, P.}
  (\bibinfo{year}{2013}).
\newblock \bibinfo{title}{{An extension of the FlowSort sorting method to deal
  with imprecision}}.
\newblock {\it \bibinfo{journal}{4OR}\/},  {\it \bibinfo{volume}{11}\/},
  \bibinfo{pages}{171--193}.
\bibitem[{Lahdelma \& Salminen(2001)}]{lahdelma-salminen2001}
\bibinfo{author}{Lahdelma, R.}, \& \bibinfo{author}{Salminen, P.}
  (\bibinfo{year}{2001}).
\newblock \bibinfo{title}{Smaa-2: Stochastic multicriteria acceptability
  analysis for group decision making}.
\newblock {\it \bibinfo{journal}{Operations Research}\/},  {\it
  \bibinfo{volume}{49}\/}, \bibinfo{pages}{444--454}.
\bibitem[{Lahdelma \& Salminen(2010)}]{Lahdelma2010}
\bibinfo{author}{Lahdelma, R.}, \& \bibinfo{author}{Salminen, P.}
  (\bibinfo{year}{2010}).
\newblock \bibinfo{title}{A method for ordinal classification in multicriteria
  decision making}.
\newblock {\it \bibinfo{journal}{International Conference on Artificial
  Intelligence and Applications}\/},  (pp. \bibinfo{pages}{420--425}).
\bibitem[{Nemery \& Lamboray(2008)}]{Nemery200890}
\bibinfo{author}{Nemery, P.}, \& \bibinfo{author}{Lamboray, C.}
  (\bibinfo{year}{2008}).
\newblock \bibinfo{title}{Flow sort: A flow-based sorting method with limiting
  or central profiles}.
\newblock {\it \bibinfo{journal}{TOP}\/},  {\it \bibinfo{volume}{16}\/},
  \bibinfo{pages}{90--113}.
\bibitem[{Pelissari et~al.(2018)Pelissari, Oliveira, Abackerli, Ben‐Amor \&
  ao}]{Pelissari2018}
\bibinfo{author}{Pelissari, R.}, \bibinfo{author}{Oliveira, M.~C.},
  \bibinfo{author}{Abackerli, A.~J.}, \bibinfo{author}{Ben‐Amor, S.}, \&
  \bibinfo{author}{ao, M. R. P.~A.} (\bibinfo{year}{2018}).
\newblock \bibinfo{title}{Techniques to model uncertain input data of
  multi‐criteria decision‐making problems: a literature review}.
\newblock {\it \bibinfo{journal}{International Transactions in Operational
  Research}\/}, . \DOIprefix\doi{https://doi.org/10.1111/itor.12598}.
\bibitem[{Pelissari et~al.(2019{\natexlab{a}})Pelissari, Oliveira, Amor \&
  Abackerli}]{PELISSARI2019}
\bibinfo{author}{Pelissari, R.}, \bibinfo{author}{Oliveira, M.~C.},
  \bibinfo{author}{Amor, S.~B.}, \& \bibinfo{author}{Abackerli, A.~J.}
  (\bibinfo{year}{2019}{\natexlab{a}}).
\newblock \bibinfo{title}{{A new FlowSort-based method to deal with information
  imperfections in sorting decision-making problems}}.
\newblock {\it \bibinfo{journal}{European Journal of Operational Research}\/},
  {\it \bibinfo{volume}{276}\/}, \bibinfo{pages}{235 -- 246}.
\bibitem[{Pelissari et~al.(2019{\natexlab{b}})Pelissari, Oliveira, Amor,
  Kandakoglu \& Helleno}]{Pelissari2019-SMAA}
\bibinfo{author}{Pelissari, R.}, \bibinfo{author}{Oliveira, M.~C.},
  \bibinfo{author}{Amor, S.~B.}, \bibinfo{author}{Kandakoglu, A.}, \&
  \bibinfo{author}{Helleno, A.~L.} (\bibinfo{year}{2019}{\natexlab{b}}).
\newblock \bibinfo{title}{{SMAA methods and their applications: a literature
  review and future research directions}}.
\newblock {\it \bibinfo{journal}{Annals of Operations Research}\/}, .
  \DOIprefix\doi{10.1007/s10479-019-03151-z}.
\bibitem[{Saaty(1994)}]{saaty1994}
\bibinfo{author}{Saaty, T.} (\bibinfo{year}{1994}).
\newblock \bibinfo{title}{How to make a decision: the analytic hierarchy
  process}.
\newblock {\it \bibinfo{journal}{European Journal of Operational Research}\/},
  {\it \bibinfo{volume}{24}\/}, \bibinfo{pages}{19--43}.
\bibitem[{Tervonen et~al.(2009)Tervonen, Figueira, Lahdelma, Dias \&
  Salminen}]{Tervonen2009a}
\bibinfo{author}{Tervonen, T.}, \bibinfo{author}{Figueira, J. R.~J.},
  \bibinfo{author}{Lahdelma, R.}, \bibinfo{author}{Dias, J. A.~J.}, \&
  \bibinfo{author}{Salminen, P.} (\bibinfo{year}{2009}).
\newblock \bibinfo{title}{{A stochastic method for robustness analysis in
  sorting problems}}.
\newblock {\it \bibinfo{journal}{European Journal of Operational Research}\/},
  {\it \bibinfo{volume}{192}\/}, \bibinfo{pages}{236--242}.
\bibitem[{Tervonen \& Lahdelma(2007)}]{Tervonen2007a}
\bibinfo{author}{Tervonen, T.}, \& \bibinfo{author}{Lahdelma, R.}
  (\bibinfo{year}{2007}).
\newblock \bibinfo{title}{{Implementing stochastic multicriteria acceptability
  analysis}}.
\newblock {\it \bibinfo{journal}{European Journal of Operational Research}\/},
  {\it \bibinfo{volume}{178}\/}, \bibinfo{pages}{500--513}.
\bibitem[{Vasto-Terrientes et~al.(2015)Vasto-Terrientes, Valls, Slowinski \&
  Zielniewicz}]{DELVASTOTERRIENTES20154910}
\bibinfo{author}{Vasto-Terrientes, L.~D.}, \bibinfo{author}{Valls, A.},
  \bibinfo{author}{Slowinski, R.}, \& \bibinfo{author}{Zielniewicz, P.}
  (\bibinfo{year}{2015}).
\newblock \bibinfo{title}{{ELECTRE-III-H: An outranking-based decision aiding
  method for hierarchically structured criteria}}.
\newblock {\it \bibinfo{journal}{Expert Systems with Applications}\/},  {\it
  \bibinfo{volume}{42}\/}, \bibinfo{pages}{4910 -- 4926}.
\bibitem[{Vetschera(2017)}]{VETSCHERA2017244}
\bibinfo{author}{Vetschera, R.} (\bibinfo{year}{2017}).
\newblock \bibinfo{title}{Deriving rankings from incomplete preference
  information: A comparison of different approaches}.
\newblock {\it \bibinfo{journal}{European Journal of Operational Research}\/},
  {\it \bibinfo{volume}{258}\/}, \bibinfo{pages}{244 -- 253}.
\bibitem[{Zopounidis \& Doumpos(2002)}]{Zopounidis2002}
\bibinfo{author}{Zopounidis, C.}, \& \bibinfo{author}{Doumpos, M.}
  (\bibinfo{year}{2002}).
\newblock \bibinfo{title}{{Multicriteria classification and sorting methods: A
  literature review}}.
\newblock {\it \bibinfo{journal}{European Journal of Operational Research}\/},
  {\it \bibinfo{volume}{138}\/}, \bibinfo{pages}{229 -- 246}.

\end{thebibliography}

\section{A numerical example of FlowSort-H}

    Let $G_{1_1}$ and $G_{2_1}$ be macro-criteria ($f=2$) with weights $w_{1_1}=0.3$ and $w_{2_1}=0.7$, respectively. Each of these macro-criteria has two sub-criteria of a second level. As there are only 2 levels of criteria, these are also the elementary criteria. The sub-criteria of $G_{1_1}$ are denoted by $g_{1_1, 1}$ e $g_{1_1, 2}$, and the sub-criteria of $G_{2_1}$ are denoted by $g_{2_1, 1}$ e $g_{2_1, 2}$. We consider that the criteria $g_{1_1, 1}, g_{2_1, 1}$ and $g_{2_1, 2}$ shall be maximized and $g_{1_1, 2}$ minimized.
        
    In this numerical decision example, two alternatives, $x_1$ and $x_2$, should be assigned to two categories $C_1$ and $C_2$. Table A.14 displays the performance values of the two alternatives according to the elementary criteria. Table A.15 presents the values of the limiting reference profiles that characterize the categories $C_1$ and $C_2$, also defined at the level of the elementary criteria.

         \begin{table}[ht!]
            \centering
             \caption{Performance values of the alternatives regarding elementary criteria in the numerical example of the FlowSort-H method.}
                \label{tab:avaliacao_exemplo_flowsort_h}
                \begin{tabular}{ccccc} \hline
                Alternativas	    &	$g_{1_1, 1}$	&	$g_{1_1, 2}$ & $g_{2_1, 1}$	&	$g_{2_1, 2}$   \\\hline
                $x_1$	            &	8	&	1	&	16	&	28	\\
                $x_2$	            &	9	&	3	&	8	&	12	\\\hline
            \end{tabular}
        \end{table}
        
        \begin{table}[ht!]
            \centering
             \caption{Limiting reference profiles for each of elementary criteria in the numerical example of the FlowSort-H method.}
                \label{tab:perfil_exemplo_flowsort_h}
                \begin{tabular}{ccccc} \hline
                Perfis de referência	    &	$g_{1_1, 1}$	&	$g_{1_1, 2}$ & $g_{2_1, 1}$	&	$g_{2_1, 2}$   \\\hline
                $r_1$	            &	10	&	0	&	20	&	30	\\
                $r_2$	            &	5	&	5	&	10	&	15	\\
                $r_3$	            &	0	&	10	&	0	&	0	\\\hline
            \end{tabular}
        \end{table}
        
        In the FlowSort-H method, weights for the criteria at all levels must be defined in such a way that the sum of the weights in a branch of the tree is not greater than 1. Thus, we define the weights of criteria $g_{1_1, 1}$ and $g_{1_1, 2}$ by $w_{1_1, 1}=0.2$ and $w_{1_1, 2}=0.8$, respectively. The weights of criteria $g_{2_1, 1}$ and $g_{2_1, 2}$ are defined by $w_{2_1, 1} = 0.4$ and $w_{2_1, 2}=0.6$. The weight of the criterion $G_{1_1}$ is $w_{1_1} = 0.3$, and the weight of criterion $G_{2_1}$ is $w_{2_1} = 0.7$. 
        
        The positive flow of the alternative $x_1$ is given by
            $$\phi^{+} (x_1) = \frac{1}{|R_1^{\mathbf{t}}| - 1} \sum_{y \in R_1-\{x_1\}} \pi(x_1, y),$$
        \noindent in which $R_1= R \cup \{x_1\} = \{r_1, r_2, r_3, x_1\}$ and, then, $|R_1| = 4$. Therefore, 
            $$\phi^{+} (x_1) = \frac{1}{3} [\pi(x_1, r_1) + \pi(x_1, r_2) + \pi(x_1, r_3)].$$
        It is now necessary to calculate the outranking degrees. To exemplify, the calculation of 
        $\pi(x_1, r_1)$ is given. Initially, it is necessary to calculate the preference functions. We choose the preference function of type 1 from the six types proposed by \cite{brans1986}. Thus, we have

        \begin{eqnarray*}
            P(d(g_{1_1, 1}(x_1), r_1)) &=& P(g_{1_1, 1}(x_1) - r_1) = P(8-10) = 0\\
            P(d(g_{1_1, 1}(x_1), r_2)) &=& P(g_{1_1, 1}(x_1) - r_2) = P(8-5) = 1\\
            P(d(g_{1_1, 1}(x_1), r_3)) &=& P(g_{1_1, 1}(x_1) - r_3) = P(8-0) = 1.
        \end{eqnarray*}
        Analogously, the preference functions of criteria $g_{2_1, 1}$ and $g_{2_1, 2}$ are calculated. As $g_{1_1, 2}$ has to be minimized, its calculation changes as follow
        \begin{eqnarray*}
            P(d(g_{1_1, 2}(x_1), r_1)) &=& P(r_1 - g_{1_1, 2}(x_1)) = P(0-1) = 0\\
            P(d(g_{1_1, 2}(x_1), r_2)) &=& P(r_2 - g_{1_1, 2}(x_1)) = P(5-1) = 1\\
            P(d(g_{1_1, 2}(x_1), r_3)) &=& P(r_3 - g_{1_1, 2}(x_1)) = P(10-1) = 1.
        \end{eqnarray*}
        Outranking degrees are then given by
        \begin{eqnarray*}
            \pi(x_1, r_1) &=& w_{1_1}[w_{1_1, 1} P(g_{1_1, 1}(x_1), r_1)+ w_{1_1, 2} P(g_{1_1, 2}(x_1), r_1)]  \\
                          &+& w_{2_1}[w_{2_1, 1} P(g_{2_1, 1}(x_1), r_1)+ w_{2_1, 2} P(g_{2_1, 2}(x_1), r_1)] \\
                          &=& 0.3(0.2 \times 0 + 0.8 \times 0) + 0.7(0.4 \times0 + 0.6\times 0) = 0\\
            \pi(x_1, r_2) &=& 0.3(0.2 \times 1 + 0.8 \times 1) + 0.7(0.4 \times1 + 0.6\times 1) = 1\\
            \pi(x_1, r_3) &=& 0.3(0.2 \times 1 + 0.8 \times 1) + 0.7(0.4 \times1 + 0.6\times 1) = 1.
        \end{eqnarray*}
        Therefore, the positive flow of the alternative $x_1$ is given by
        $$\phi^{+}(R_1) = \frac{1}{3}(0+1+1) = \frac{2}{3} = 0.667.$$
        In an analogous way, the positive flows of the profiles related to the alternative $x_1$ are obtained: $\phi^{+}_{R_1}(r_1)=1$, $\phi^{+}_{x_1}(r_2)=0.333$ e $\phi^{+}_{x_1}(r_3)=0$. Therefore, 
        $\phi^{+}_{R_1}(r_1) > \phi^{+}(x_1) > \phi^{+}_{R_1}(r_2)$, which implies that alternative $x_1$ must be assigned to the $C_1$ category.
        
        Following the same line of calculation, we obtain the positive, negative and net flows of alternatives $x_1$ and $x_2$ and of the reference profiles, presented in Tables A.16 and A.17, respectively.
        
        \begin{table}[ht!]
            \small
            \centering
             \caption{Flows of alternatives in the numerical example of the FlowSort-H method.}
                \label{tab_ap:fluxo_alte_hier}
                \begin{tabular}{cccc} \hline
                Alternativas    &	$\phi^{+}(.)$	& $\phi^{-}(.)$ & $\phi(.)$ \\\hline
                 $x_1$          &	0.667	      &	0.333      &     0.333  	\\
                 $x_2$          &	0.433         &	0.566     &     -0.133 	\\\hline
            \end{tabular}
        \end{table}
        
        \begin{table}[ht!]
            \small
            \centering
             \caption{Flows of limiting profiles in the numerical example of the FlowSort-H method.}
                \label{tab_ap:fluxo_perfil_hier}
                \begin{tabular}{ccccccc} \hline
                Alternativas    &	$\phi^{+}_{R_1}(.)$	& $\phi^{-}_{R_1}(.)$ & $\phi_{R_1}(.)$ &	$\phi^{+}_{R_2}(.)$	& $\phi^{-}_{R_2}(.)$ & $\phi_{R_2}(.)$\\\hline
                 $r_1$          &	1	            &	0        	&     1  	 & 1         &  0     & 1 \\
                 $r_2$          &	0.333	        &	0.666       &    -0.333  & 0.566	 & 0.433  &  0.133\\
                 $r_3$          &	0	            &	1        	&     -1 	 &  0        &   1    & -1	\\\hline
            \end{tabular}
        \end{table}
        
        Finally, based on the assignment rules defined in \eqref{atribuicao_pos_flow}, \eqref{atribuicao_neg_flow}  and \eqref{atribuicao_net_flow}, we obtain:

        \begin{eqnarray*}
           &\mbox{ Since }& \phi^{+}_{R_1} (r_1) = 1 \geq  \phi^{+} (x_1) = 0.667 > \phi^{+}_{R_1} (r_{2}) = 0.333, \mbox{ then } C_{\phi^{+}}(x_1) = C_1;\\
           &\mbox{ since }& \phi^{-}_{R_1} (r_1) = 0 < \phi^{-}(x_1) = 0.333 \leq \phi^{-}_{R_1} (r_{2}) = 0.666, \mbox{ then } C_{\phi^{-}}(x_1) = C_1;\\
           &\mbox{ since }& \phi_{R_1} (r_1) = 1 \geq \phi (x_1) =0.334 > \phi_{R_1} (r_2) = -0.333, \mbox{ then } C_{\phi}(x_1) = C_1.
        \end{eqnarray*}
        
        Thus, using any of the assignment rules, alternative $x_1$ is assigned to category $C_1$. In the same way, we conclude that, by applying any of the assignment rules, alternative $x_2$ is assigned to category $C_2$.
        
        To know in which criteria alternative $x_2$ must improve to be assigned to category $C_1$, we must calculate the single-criteria flows of $x_2$ related to criteria $g_{1_1}$ and $g_{2_1}$ and their sub-criteria.
        
        \begin{eqnarray*}
            \phi_{g_{1_1}}^{+}(x_2) &=& \frac{1}{|R_2| - 1} \left(\sum_{h=1}^{3} \pi_{g_{1_1, 1}}(x_2, r_h) +\sum_{h=1}^{3} \pi_{g_{1_1, 2}}(x_2, r_h)\right)\\
                                    &=& \frac{1}{3} (\pi_{g_{1_1, 1}}(x_2, r_1) + \pi_{g_{1_1, 1}}(x_2, r_2) + \pi_{g_{1_1, 1}}(x_2, r_3))
                                    +\pi_{g_{1_1, 2}}(x_2, r_1) + \pi_{g_{1_1, 2}}(x_2, r_2) + \pi_{g_{1_1, 2}}(x_2, r_3))\\
                                    &=& \frac{1}{3}~ (w_{1_1, 1}(~P(g_{1_1, 1}(x_2) - r_1) + P(g_{1_1, 1}(x_2) - r_2) + P(g_{1_1, 1}(x_2) - r_3)) \\
                                    &~& + ~ w_{1_1, 2}(~P(g_{1_1, 2}(x_2) - r_1) + P(g_{1_1, 2}(x_2) - r_2) + P(g_{1_1, 2}(x_2) - r_3)))\\
                                    &=& \frac{1}{3}~ (0.2\times(0 + 1 + 1) + 0.8\times(0 + 1 + 1))\\
                                    &=& \frac{1}{3}~ (0.2\times 2 + 0.8\times 2) = 0.667
          \end{eqnarray*}     
          
          \begin{eqnarray*}
            \phi_{g_{2_1}}^{+}(x_2) &=& \frac{1}{|R_2| - 1} \left(\sum_{h=1}^{3} \pi_{g_{2_1, 1}}(x_2, r_h) +\sum_{h=1}^{3}\pi_{g_{2_1, 2}}(x_2, r_h)\right)\\
                                    &=& \frac{1}{3} ((\pi_{g_{2_1, 1}}(x_2, r_1) + \pi_{g_{2_1, 1}}(x_2, r_2) + \pi_{g_{2_1, 1}}(x_2, r_3)) \\
                                    &~& + ~ (\pi_{g_{2_1, 2}}(x_2, r_1) + \pi_{g_{2_1, 2}}(x_2, r_2) + \pi_{g_{2_1, 2}}(x_2, r_3)))\\
                                    &=& \frac{1}{3}~ (w_{2_1, 1}(~P(d(g_{2_1, 1}(x_2) - r_1)) + P(d(g_{2_1, 1}(x_2) - r_2)) + P(d(g_{2_1, 1}(x_2) - r_3))) \\
                                    &~& + ~ w_{2_1, 2}(~P(d(g_{2_1, 2}(x_2) - r_1)) + P(d(g_{2_1, 2}(x_2) - r_2)) + P(d(g_{2_1, 2}(x_2) - r_3))))\\
                                    &=& \frac{1}{3}~ (0.4(0 + 0 + 1) + 0.6(0 + 0 + 1))\\
                                    &=& \frac{1}{3}~ (0.4(1) + 0.6(1)) = 0.333\\
        \phi_{g_{1_1, 1}}^{+}(x_2) &=& \frac{1}{|R_2| - 1} \left(\sum_{h=1}^{3} \pi_{g_{1_1, 1}}(x_2, r_h)\right)\\
                                    &=& \frac{1}{3} (\pi_{g_{1_1, 1}}(x_2, r_1) + \pi_{g_{1_1, 1}}(x_2, r_2) + \pi_{g_{1_1, 1}}(x_2, r_3)) \\
                                    &=& \frac{1}{3}~( P(d(g_{1_1, 1}(x_2) - r_1)) + P(d(g_{1_1, 1}(x_2) - r_2)) + P(d(g_{1_1, 1}(x_2) - r_3))) \\
                                    &=& \frac{1}{3}~ (0 + 1 + 1) = 0.667\\
        \phi_{g_{1_1, 2}}^{+}(x_2) &=& \frac{1}{|R_2| - 1} \left(\sum_{h=1}^{3} \pi_{g_{1_1, 2}}(x_2, r_h)\right)\\
                                    &=& \frac{1}{3} (\pi_{g_{1_1, 2}}(x_2, r_1) + \pi_{g_{1_1, 2}}(x_2, r_2) + \pi_{g_{1_1, 2}}(x_2, r_3)) \\
                                    &=& \frac{1}{3}~( P(d(g_{1_1, 2}(x_2) - r_1)) + P(d(g_{1_1, 2}(x_2) - r_2)) + P(d(g_{1_1, 2}(x_2) - r_3))) \\
                                    &=& \frac{1}{3}~ (0 + 1 + 1) = 0.667\\
        \phi_{g_{2_1, 1}}^{+}(x_2) &=& \frac{1}{|R_2| - 1} \left(\sum_{h=1}^{3} \pi_{g_{2_1, 1}}(x_2, r_h)\right)\\
                                    &=& \frac{1}{3} (\pi_{g_{2_1, 1}}(x_2, r_1) + \pi_{g_{2_1, 1}}(x_2, r_2) + \pi_{g_{2_1, 1}}(x_2, r_3)) \\
                                    &=& \frac{1}{3}~( P(d(g_{1_2, 1}(x_2) - r_1)) + P(d(g_{1_2, 1}(x_2) - r_2)) + P(d(g_{1_2, 1}(x_2) - r_3))) \\
                                    &=& \frac{1}{3}~ (0 + 0 + 1) = 0.333\\
        \phi_{g_{2_1, 2}}^{+}(x_2) &=& \frac{1}{|R_2| - 1} \left(\sum_{h=1}^{3} \pi_{g_{2_1, 2}}(x_2, r_h)\right)\\
                                    &=& \frac{1}{3} (\pi_{g_{2_1, 2}}(x_2, r_1) + \pi_{g_{2_1, 2}}(x_2, r_2) + \pi_{g_{2_1, 2}}(x_2, r_3)) \\
                                    &=& \frac{1}{3}~( P(d(g_{2_1, 2}(x_2) - r_1)) + P(d(g_{2_1, 2}(x_2) - r_2)) + P(d(g_{2_1, 2}(x_2) - r_3))) \\
                                    &=& \frac{1}{3}~ (0 + 0 + 1) = 0.333\\
        \end{eqnarray*}
        
        The values of the single-criterion flows of alternative $x_2$ calculated above are shown in Table \ref{tab:fluxo_alt_uni}. Following the same line of calculation, the values of the single-criterion flows of the reference profiles are obtained. The values are shown in Table \ref{tab:fluxo_perfis_uni}.
        
        \begin{table}[ht!]
            \small
            \centering
             \caption{Single-criterion flow of the alternative $x_2$ related to the numerical example of FlowSort-H.}
                \label{tab:fluxo_alt_uni}
                \begin{tabular}{ccccccc} \hline
                Alternativa    & $\phi_{g_{1_1}}(.)$ & $\phi_{g_{2_1}}(.)$ & $\phi_{g_{1_1, 1}}(.)$	& $\phi_{g_{1_1, 2}}(.)$ & $\phi_{g_{2_1, 1}}(.)$ & $\phi_{g_{2_1, 2}} (.)$ \\\hline
                 $x_2$          &  0.667     & 0.333   &    0.667   & 0.667  & 0.333 & 0.333\\\hline
            \end{tabular}
        \end{table}
    
        \begin{table}[ht!]
            \small
            \centering
             \caption{Single-criterion flow of the limiting profiles of the numerical example of FlowSort-H.}
                \label{tab:fluxo_perfis_uni}
                \begin{tabular}{ccccccc} \hline
                Alternativa/ Perfil    & $\phi_{R_2, g_{1_1}}(.)$ & $\phi_{R_2, g_{2_1}}(.)$ & $\phi_{R_2, g_{1_1, 1}}(.)$	& $\phi_{R_2, g_{1_1, 2}}(.)$ & $\phi_{R_2, g_{2_1, 1}}(.)$ & $\phi_{R_2, g_{2_1, 2}} (.)$ \\\hline
                 $x_2$          &  0.667     & 0.333   &    0.667   & 0.667  & 0.333 & 0.333\\\hline
                 $r_1$          &	1	      &	1      &     1      & 1      & 1     & 1  	\\
                 $r_2$          &	- 0.333	  &	0.333  &     -0.333 & -0.333 & 0.333 & 0.333  	\\
                 $r_3$          &	-1	      &	-1     &     -1    & -1      & -1    & -1   	\\\hline
            \end{tabular}
        \end{table}

       Thus, $\phi_{R_2, g_{1_1}}(r_1) = 1 \geq \phi (x_2) =0.667 > \phi_{R_2, g_{1_1}}(r_2) = - 0.333$. Therefore, $C_{\phi}(x_2) = C_1$ with respect criterion $g_{1_1}$. Analogously, $\phi_{R_2, g_{2_1}}(r_2) = 0.333 \geq \phi (x_2) =0.333 > \phi_{R_2, g_{2_1}}(r_3) = - 1$, and therefore $C_{\phi}(x_2) = C_2$ when the alternative $x_2$ is analyzed with respect to only criterion $g_{2_1}$.
        
        Analyzing the sub-criteria of the criterion $g_{2_1}$, we have $C_{\phi}(x_2) = C_2$ for both sub-criteria $g_{2_1, 1}$ and $g_{2_1, 2}$. Therefore, improving the performance of alternative $x_2$ regarding criteria $g_{2_1, 1}$ and $g_{2_1, 2}$ shall result, at some point of this modification, in an assignment of the alternative $x_2$ to category $C_1$.

\end{document}